%% file: arxiv.tex
\documentclass[letterpaper]{article}
\usepackage[preprint]{aaai2027}
\usepackage[hyphens]{url}
\usepackage{graphicx}
\usepackage{natbib}
\usepackage{caption}
\usepackage{amsmath}
\usepackage{amssymb}
\usepackage{booktabs}
\usepackage{array}
\usepackage{algorithm}
\usepackage{algorithmic}
\usepackage{listings}

\lstdefinestyle{appendixcode}{
  basicstyle=\footnotesize\ttfamily,
  numbers=none,
  showstringspaces=false,
  keepspaces=true,
  breaklines=true,
  columns=fullflexible,
  frame=single,
  framerule=0.3pt,
  tabsize=2
}
\lstdefinestyle{appendixprompt}{
  style=appendixcode,
  frame=single
}
\newcommand{\method}{CARE}
\newcommand{\olympusdataset}{Olympus Suzuki Coupling (i)}
\newcommand{\chemlexdataset}{ChemLex carboxylic acid and amine coupling}
\newcommand{\olympus}{Olympus Suzuki (i)}
\newcommand{\chemlex}{ChemLex}
\newcommand{\bhsuite}{Buchwald--Hartwig A--E}
\newcommand{\directarylation}{Direct Arylation}
\newcommand{\smiles}[1]{\path{#1}}

\title{CARE: Context-Aware Ranking Evolution with Executable Scoring Programs for Budgeted Reaction Optimization}
\author{
Guanyu Liu\textsuperscript{1}\equalcontrib,
Weiyi Kong\textsuperscript{2}\equalcontrib,
Chao Tang\textsuperscript{1}\equalcontrib,
Zeyu Wang\textsuperscript{3},
Boer Zhang\textsuperscript{4},
Baiqing Li\textsuperscript{5},
Peiyu Zhang\textsuperscript{5,$\dagger$},
Tianyu Shi\textsuperscript{6,$\dagger$}
}
\affiliations{
\textsuperscript{1}University of Macau,
\textsuperscript{2}University of Toronto,
\textsuperscript{3}UCLA,
\textsuperscript{4}Harvard University,
\textsuperscript{5}XtalPi,
\textsuperscript{6}McGill University\\
\texttt{peiyu.zhang@xtalpi.com},
\texttt{tianyu.shi3@mcgill.ca}\\
$\dagger$ Corresponding authors.
}

\begin{document}

\maketitle

\begin{abstract}
High-throughput experimentation can evaluate many reaction conditions, yet combinatorial condition spaces still exceed the available experiment budget. This makes experiment selection a sequential decision problem: each new condition must be chosen from limited observations before its outcome is known. LLMs can express task-specific selection logic. An LLM can suggest the next experiment directly, but such a one-off recommendation does not provide a reusable scoring rule that can be applied to all remaining conditions, systematically checked, and updated after new results. We introduce CARE, in which the LLM writes a reusable rule for ranking the untested conditions but does not select the experiment directly. A non-LLM reference policy provides the default condition, and an intervention gate decides whether to replace it with the program-derived alternative. CARE records this decision before the selected outcome is revealed. Each new outcome updates controller state and can trigger reuse, modification, or replacement of the active program without updating the LLM parameters. In matched offline replay with 30 seeds on eight reaction-optimization tasks, CARE attains the lowest normalized regret, the highest normalized best-so-far AUC, and the highest Top-1\% Success@15 among the evaluated methods. These results support using a scoring program written by an LLM as one component of a reference-conditioned optimizer rather than as a standalone experiment selector.
\end{abstract}

\input{sections/introduction}
\input{sections/related_work}
\input{sections/problem_setting}
\input{sections/method}
\input{sections/experiments}
\input{sections/conclusion}
\input{sections/limitations}

\clearpage
\bibliography{custom}

\clearpage
\input{sections/appendix}

\end{document}

%% file: sections/introduction.tex
\section{Introduction}
\label{sec:introduction}

Optimizing a chemical reaction requires choosing among combinations of catalysts, solvents, temperatures, substrates, and other conditions. These combinations grow quickly, often beyond the number of experiments a laboratory can run. High-throughput experimentation (HTE) raises throughput through automated, miniaturized, and parallel experiments \citep{tom2024selfdrivinglabs,sin2025parallel}, but it does not remove the need to allocate a limited budget across a much larger design space.

We study this allocation problem when the candidate conditions are enumerated in advance. A controller begins with a small set of observed outcomes, selects one unevaluated condition, and updates its strategy after the corresponding yield or conversion is revealed. We compare controllers through offline replay on completed chemistry benchmarks. The benchmark provides the outcomes, but each outcome remains hidden until the controller commits to its next condition. This setting preserves the sequential feedback of an experimental campaign while enabling matched comparisons under the same initial observations and budget.

Bayesian optimization (BO) is a standard approach to this problem. It fits a surrogate model to the revealed outcomes and uses an acquisition function to rank the remaining conditions. Different acquisition rules encode different preferences for predicted performance, uncertainty, and exploration, and modern BO systems support continuous, categorical, and mixed condition spaces \citep{botorch2019,smac2011,tpe2011,gryffin2021,shields2021edbo,baybe2025,bofire2024}. Such numerical models provide a useful anchor when observations are sparse.

LLMs offer a complementary source of selection logic. Scientific agents use them to plan procedures and coordinate tools \citep{boiko2023coscientist,chemcrow2024,ruan2024llmrdf,song2025chemagents}, while BO systems use LLMs for priors, predictions, acquisition strategies, or candidate proposals \citep{llmbo2024,cisse2025bora,cisse2026bora,boicl2026,patel2026distilling}. Studies of scientific optimization also show why the LLM's role matters: direct agents may use numerical feedback inconsistently, whereas hybrid designs can preserve the acquisition process of a statistical model \citep{gupta2025llmbo,soberlook2024}. This motivates a concrete question: can an LLM adapt an executable scoring rule from revealed outcomes while a non-LLM policy anchors the final experiment choice?

CARE answers this question by treating each round as a reference-conditioned decision. Given the experimental results observed so far and the remaining untested conditions, an ensemble-based non-LLM reference policy aggregates several conventional numerical rankers to produce a reference candidate---the default choice---and a numerical support summary. In parallel, an LLM updates an executable \emph{scoring program}, whose maximum forms an initial proposal. CARE then applies exploitation or exploration refinement to form an alternative and uses an intervention gate to select between the reference and alternative. Figure~\ref{fig:care_pipeline} summarizes this decision process.

New experimental results guide the controller to reuse, modify, or replace the scoring program, which is validated before use; if validation fails, CARE defaults to the numerical reference. This process updates the scoring logic without giving the LLM direct control over experiment selection. The gate selects between the reference and alternative candidates, and the resulting decision record is carried forward to the next round. This separation distinguishes program adaptation from final experiment selection.

Across 30 matched seeds on each of eight reaction-optimization tasks, \method{} attains the lowest normalized regret, the highest normalized best-so-far AUC, and the highest Top-1\% Success@15 among the evaluated methods. These metrics respectively capture final condition quality, early-search efficiency, and elite-condition discovery within the 15-experiment budget. The suite spans an emulated Suzuki coupling benchmark and HTE data from carboxylic acid--amine coupling, Buchwald--Hartwig amination, and direct arylation. On the two tasks with complete paired program-selection controls, all four continuous-metric comparisons with the reference policy remain significant after Holm correction. Component controls on three representative HTE tasks show that direct program selection and fixed scoring logic underperform CARE, and that removing planner adaptation or either refinement mode weakens the complete controller.

\paragraph{Contributions.}
% CARE contributes a stateful, reference-conditioned controller that factorizes each selection into numerical reference construction, validated program update, dual-mode proposal refinement, and candidate comparison. This formulation places an evolving scoring program in a selective proposal role rather than using it as a direct experiment selector. Matched replay across eight tasks shows lower normalized regret, higher normalized BSF-AUC, and higher Top-1\% campaign success than the evaluated BO baselines; direct-selection controls and component ablations isolate the contribution of the controller design under the same experiment budget.

CARE contributes a reference-conditioned controller that carries information across rounds and separates each selection into four components: selecting a reference candidate using conventional numerical rankers, updating and validating an executable scoring program, refining the program-derived proposal through exploitation or exploration, and comparing the resulting alternative with the reference. Under this design, the evolving scoring program proposes alternatives for consideration rather than selecting experiments directly. Across eight tasks, matched replay shows lower normalized regret, higher normalized BSF-AUC, and higher Top-1\% campaign success than the evaluated BO baselines; direct-selection controls and component ablations, conducted under the same experiment budget, isolate the contribution of the controller design.

%% file: sections/related_work.tex
\section{Related Work}
\label{sec:related-work}

\paragraph{Reaction condition optimization and HTE.}
HTE evaluates arrays of reaction conditions whose outcomes are measured by quantities such as yield, conversion, or selectivity. Platforms such as phactor standardize designs and results across plates containing 24 to 1{,}536 wells \citep{mahjour2023phactor}, but the full combination of catalysts, ligands, solvents, temperatures, and substrates can still exceed experimental capacity. BO allocates this limited budget by fitting a surrogate to observed outcomes and selecting new conditions with an acquisition function \citep{shields2021edbo,desimpel2026bo}. Recent work extends this approach to parallel HTE batches and the search for conditions that perform well across multiple substrates \citep{sin2025parallel,wang2024bandit}. CARE sequentially selects from a predefined set of reaction conditions and observes each outcome only after choosing the corresponding condition.

\paragraph{LLMs in chemistry and scientific optimization.}
ChemCrow equips a general model with chemistry tools \citep{chemcrow2024}, while Coscientist and an automated synthesis platform connect language model planning to laboratory operations \citep{boiko2023coscientist,ruan2024llmrdf}. Chemma instead fine-tunes a chemistry model for tasks that include yield prediction and uses those predictions in BO and active learning \citep{zhang2025chemma}. These systems use an LLM to plan, operate tools, or predict chemical outcomes. CARE targets a different role: given an enumerated condition set, the LLM writes the scoring logic used to rank candidates during sequential optimization.

\paragraph{LLM roles in BO.}
LLM-assisted BO methods differ mainly in where the language model enters the optimization loop. LLAMBO uses an LLM for warm starting, surrogate modeling, and candidate sampling \citep{llmbo2024}. LMABO chooses an acquisition function from a portfolio based on the current BO state \citep{ngo2026lmabo}, whereas BORA uses an LLM to interpret the trajectory and suggest promising search regions within a hybrid optimizer \citep{cisse2025bora}. Evaluations on molecular and scientific tasks indicate a useful division of labor: LLM representations or priors can help under suitable training, while statistical models remain valuable for incorporating numerical feedback and directing acquisition \citep{soberlook2024,gupta2025llmbo}. CARE follows this hybrid view but gives the LLM a distinct role as the author of an explicit scoring program. The reference policy supplies the default candidate, and the intervention gate completes selection by comparing the constructed candidates.

\paragraph{Program evolution with LLMs.}
FunSearch couples programs generated by an LLM with an external evaluator and searches a population of scored programs \citep{funsearch2023}. EoH evolves both natural-language heuristic ideas and their code, while HSEvo uses population diversity to guide evolutionary program search \citep{liu2024eoh,dat2025hsevo}. Related systems apply executable code generation to reward functions and BO acquisition functions \citep{eureka2023,funbo2025}. This representation makes the generated logic runnable and inspectable, while an external procedure determines how it is evaluated or deployed. CARE retains these representational advantages but maintains a single active scoring program throughout each campaign, using new experimental results to decide whether to reuse, modify, or replace it.

%% file: sections/problem_setting.tex
\section{Problem Setting and Overview}
\label{sec:problem}

\paragraph{Finite candidate optimization and replay protocol.}
Let $\mathcal{X}=\{x_i\}_{i=1}^{N}$ be a finite set of candidate experimental conditions. Each $x_i$ contains only attributes that are available before an experiment is selected. A hidden outcome map $f:\mathcal{X}\rightarrow\mathbb{R}$ assigns an outcome to each condition, with larger values preferred. Each replay begins with a set $\mathcal{I}_0\subset\mathcal{X}$ of five evaluated conditions. The history and remaining candidate set before round 1 are
\[
\mathcal{H}_1=\{(x,f(x)):x\in\mathcal{I}_0\},
\qquad
\mathcal{C}_1=\mathcal{X}\setminus\mathcal{I}_0.
\]
At round $t$, a controller observes $\mathcal{H}_t$ and the permitted attributes of $\mathcal{C}_t$, then selects $a_t\in\mathcal{C}_t$. The evaluator reveals $f(a_t)$ only after this decision and updates
\[
\mathcal{H}_{t+1}=\mathcal{H}_t\cup\{(a_t,f(a_t))\},
\qquad
\mathcal{C}_{t+1}=\mathcal{C}_t\setminus\{a_t\}.
\]
The task is budgeted sequential optimization over $\mathcal{X}$. Offline replay supplies the outcomes, but the controller cannot propose conditions outside the candidate set.

\paragraph{Evaluation measures.}
Our experiments allow $B=10$ selections after the five initial observations. The best outcome available after round $t$ is
\[
b_t=\max\{y:(x,y)\in\mathcal{H}_{t+1}\}.
\]
For a decision made before round $t$, we write
\[
\bar b_t=\max\{y:(x,y)\in\mathcal{H}_t\},
\]
so that $b_t=\bar b_{t+1}$. This distinction keeps the outcome selected in round $t$ out of that round's decision.
We normalize the terminal incumbent and best-so-far trajectory using the complete-pool range after replay.
Let $y^- = \min_{x\in\mathcal X}f(x)$ and $y^\star=\max_{x\in\mathcal X}f(x)$ denote the complete-pool extrema, used only after a replay is complete, and let
\[
q_t=\frac{b_t-y^-}{y^\star-y^-}.
\]
We report normalized regret after the full 15-condition campaign,
\[
\mathrm{NR}@15=100(1-q_B),
\]
and normalized best-so-far area under the adaptive trajectory,
\[
\mathrm{nBSF\text{-}AUC}@10=\frac{100}{B}\sum_{t=1}^{B}q_t.
\]
The former measures the remaining distance to the pool optimum, while the latter rewards reaching strong incumbents early. We also report whether the campaign finds a member of the top 1\% of the complete pool. If $\mathcal E_{1\%}$ is the set of the $\lceil 0.01N\rceil$ highest-outcome conditions, then
\[
\mathrm{Success}@15=\mathbf{1}\!\left[\bigl(\mathcal I_0\cup\{a_t\}_{t=1}^{B}\bigr)\cap\mathcal E_{1\%}\neq\varnothing\right].
\]
Thus, ``@15'' counts the five initial and ten selected conditions, whereas ``@10'' averages the ten post-selection incumbents. Pool extrema and elite ranks are evaluation-only quantities.

\paragraph{Decision-time interface and state transition.}
At the start of round $t$, the controller receives the decision-time view
\[
\mathcal{V}_t=(\mathcal{H}_t,\mathcal{C}_t,s_t),
\]
where $s_t$ is persistent controller state derived from earlier rounds, such as active-program metadata, refinement usage, and prior decision records. Every pre-outcome CARE component is a function of $\mathcal{V}_t$: it may use revealed outcomes in $\mathcal{H}_t$ and permitted attributes in $\mathcal{C}_t$, but not the hidden outcomes of the remaining candidates. Before observation, CARE writes a decision record $\ell_t$ containing the constructed candidates and the selection rationale. Once the evaluator reveals $y_t=f(a_t)$, the controller updates its persistent state as
\[
s_{t+1}=\mathcal{U}(s_t,\ell_t,(a_t,y_t)).
\]
Thus, outcome feedback enters the next decision through $\mathcal{H}_{t+1}$ and $s_{t+1}$, not through the current selection.

\paragraph{Reference, alternative, and selected candidates.}
CARE factorizes each selection into candidate construction and reference-conditioned selection. A \emph{non-LLM reference policy}, hereafter the \emph{reference policy}, produces a reference candidate $a_t^{\mathrm{ref}}$ and a numerical reference summary. A validated scoring program maps $\mathcal{V}_t$ to one finite score for each remaining candidate,
\[
\pi_t:\mathcal{V}_t
\longmapsto \bigl(z_t(x)\bigr)_{x\in\mathcal{C}_t}
\in\mathbb{R}^{|\mathcal{C}_t|},
\]
whose maximum is the initial program proposal. Constrained proposal refinement may then form an alternative candidate $a_t^{\mathrm{alt}}$. A reference-conditioned intervention gate compares the two constructed candidates and selects
\[
a_t\in\{a_t^{\mathrm{ref}},a_t^{\mathrm{alt}}\}.
\]
When no valid program is available, the alternative is absent and CARE selects $a_t^{\mathrm{ref}}$. Figure~\ref{fig:care_pipeline} shows this order; Section~\ref{sec:method} formalizes the four controller components.

%% file: sections/method.tex
\section{Method}
\label{sec:method}

\begin{figure*}[tb]
\centering
\includegraphics[width=0.98\linewidth]{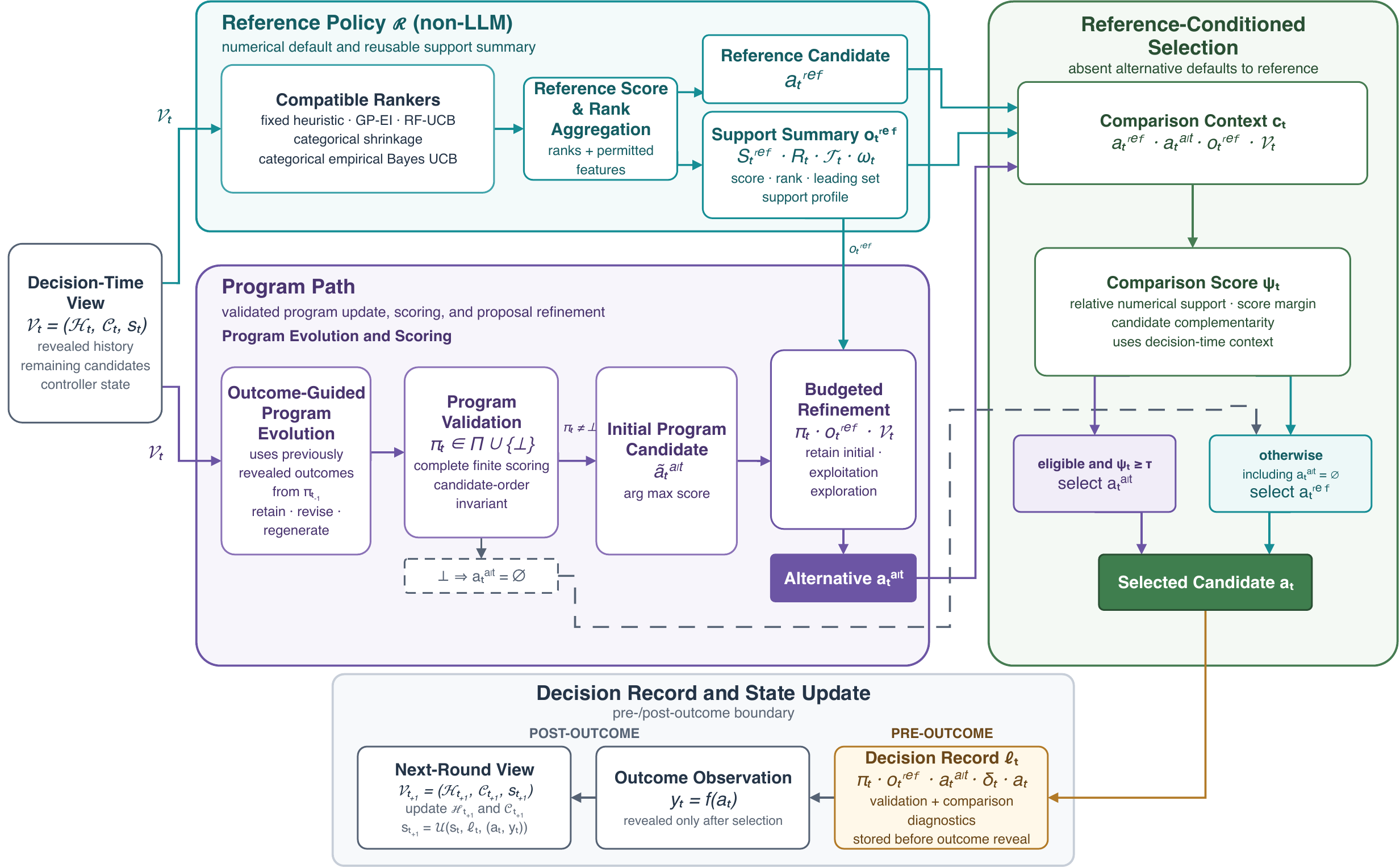}
\caption{One round of CARE. From the decision-time view, the reference path produces a numerical candidate and support summary, while the program path produces a validated scoring program and refined alternative. Reference-conditioned selection chooses between these candidates; the observed outcome then updates the next-round state.}
\label{fig:care_pipeline}
\end{figure*}

CARE is a stateful, reference-conditioned controller. Given the decision-time view $\mathcal{V}_t$, it constructs a numerical reference, evolves an executable scoring program from the revealed trajectory, refines the resulting program candidate when warranted, and selects between the two candidates. This separation prevents the adaptive program from replacing the numerical reference without an explicit, reference-conditioned comparison. Formally, one round is
\begin{equation}
\begin{aligned}
\bigl(o_t^{\mathrm{ref}},a_t^{\mathrm{ref}}\bigr)
&=\mathcal{R}(\mathcal{V}_t), \\
\pi_t
&=\mathcal{P}(\pi_{t-1},\mathcal{V}_t), \\
a_t^{\mathrm{alt}}
&=\operatorname{Refine}(\pi_t,o_t^{\mathrm{ref}},\mathcal{V}_t), \\
a_t
&=\operatorname{Select}(a_t^{\mathrm{ref}},a_t^{\mathrm{alt}},o_t^{\mathrm{ref}},\mathcal{V}_t).
\end{aligned}
\label{eq:care_factorization}
\end{equation}
Here $\mathcal{R}$ constructs the reference candidate and its numerical support summary, whereas $\mathcal{P}$ maps the previous program and currently revealed information to a valid scoring program or $\bot$. When a valid program is available, $\operatorname{Refine}$ starts from its maximum-scoring candidate; otherwise it returns $\varnothing$. $\operatorname{Select}$ then retains the reference whenever the alternative is absent or does not meet its comparison criterion. Figure~\ref{fig:care_pipeline} and Table~\ref{tab:care_components} summarize these interfaces.

\begin{table*}[tb]
\centering
\caption{Functional roles of the four CARE components. The supplementary material specifies their reported implementations.}
\label{tab:care_components}
\small
\setlength{\tabcolsep}{4pt}
\begin{tabular}{@{}p{0.16\textwidth}p{0.25\textwidth}p{0.39\textwidth}p{0.14\textwidth}@{}}
\toprule
Component & Input at decision time & Role & Output \\
\midrule
Reference policy & $\mathcal{V}_t$ & Aggregates compatible numerical rankers into a candidate and shared support summary & $(o_t^{\mathrm{ref}},a_t^{\mathrm{ref}})$ \\
Program evolution and scoring & $\mathcal{V}_t$ and $\pi_{t-1}$ & Evolves and validates a scoring program, then ranks the remaining candidate pool & $\pi_t$ and $\widetilde a_t^{\mathrm{alt}}$ \\
Budgeted refinement & $\mathcal{V}_t$, $\pi_t$, and $o_t^{\mathrm{ref}}$ & Optionally replaces the initial program candidate through exploitation or exploration & $a_t^{\mathrm{alt}}$ \\
Reference-conditioned selection & $\mathcal{V}_t$, $o_t^{\mathrm{ref}}$, and the two candidates & Compares the constructed candidates and selects one of them & $a_t$ and decision record $\ell_t$ \\
\bottomrule
\end{tabular}
\end{table*}

\subsection{Reference Policy}
\label{sec:reference_proposal}

The reference policy supplies a numerical default and a reusable support summary for the downstream controller. Given $\mathcal{V}_t$, it instantiates the rankers compatible with the available candidate attributes. Let $r_{t,e}(x)$ be the rank assigned to candidate $x$ by ranker $e\in\mathcal{E}_t$, and let $S_t^{\mathrm{ref}}(x)$ denote the aggregate score obtained from these ranks and the decision-time reference features $\phi_t^{\mathrm{ref}}(x)$. Sorting $S_t^{\mathrm{ref}}$ over $\mathcal{C}_t$ gives the aggregate rank $R_t(x)$ and a leading reference set $\mathcal{T}_t$. We also retain a support profile $\omega_t(x)$ recording which individual rankers place $x$ in their leading sets. Thus,
\begin{equation}
\begin{aligned}
o_t^{\mathrm{ref}}
&=\bigl(S_t^{\mathrm{ref}},R_t,\mathcal{T}_t,\omega_t\bigr),\\
a_t^{\mathrm{ref}}
&=\operatorname*{arg\,max}_{x\in\mathcal{C}_t}S_t^{\mathrm{ref}}(x),\\
\bigl(o_t^{\mathrm{ref}},a_t^{\mathrm{ref}}\bigr)
&=\mathcal{R}(\mathcal{V}_t).
\end{aligned}
\label{eq:reference_support_set}
\end{equation}
The reported reference instantiation combines attribute, surrogate, and categorical rankers when their inputs are available. Its rank transforms, schedules, support-set size, and feature definitions are specified in the supplementary material. In the main controller, $o_t^{\mathrm{ref}}$ is simply the numerical context used to refine and compare a program candidate.

\subsection{Outcome-Guided Program Evolution}
\label{sec:program_evolution}

The scoring program adapts as feedback accumulates, but its role is to revise a ranking rule rather than select an experiment directly. Let $\Pi$ denote the class of executable scoring programs. At round $t$,
\begin{equation}
\pi_t=\mathcal{P}(\pi_{t-1},\mathcal{V}_t)\in\Pi\cup\{\bot\}.
\label{eq:program_update}
\end{equation}
The map $\mathcal{P}$ may retain, revise, or regenerate the previous program. Any newly produced program must satisfy a fixed executable-scoring interface. If an update fails, $\mathcal{P}$ follows the recorded fallback policy: it may retain a prior valid program or return $\bot$. When $\pi_t=\bot$, CARE falls back to the reference policy. Thus, outcome feedback can change the ranking rule while preserving a well-defined controller interface.

When $\pi_t\neq\bot$, let $z_t(x)=\pi_t(\mathcal{V}_t)(x)$ be its score for candidate $x$. Its initial program candidate is
\begin{equation}
\widetilde a_t^{\mathrm{alt}}
=\operatorname*{arg\,max}_{x\in\mathcal{C}_t}z_t(x).
\label{eq:alternative_proposal}
\end{equation}
The supplementary material specifies the prompts, validation tests, and fallback implementation. The main text uses only the resulting contract: a valid program maps $\mathcal{V}_t$ to a complete finite ranking of $\mathcal{C}_t$.

\subsection{Budgeted Refinement}
\label{sec:proposal_refinement}

The program maximum is a useful starting point, but it need not be the candidate that most usefully complements the numerical reference. CARE therefore applies a budgeted refinement stage with $\mathcal{M}=\{\mathrm{exploit},\mathrm{explore}\}$. Each mode $m\in\mathcal{M}$ is specified by an activation predicate $g_m$, a feasible set $\mathcal{A}_{m,t}\subseteq\mathcal{C}_t$, a refinement score $h_{m,t}$, and a budget $B_m$. These quantities depend only on the decision-time context $\xi_t=(\pi_t,o_t^{\mathrm{ref}},\mathcal{V}_t)$. If $n_{m,t}$ is the number of earlier uses of mode $m$, the available modes are
\begin{equation}
\mathcal{M}_t
=\left\{m\in\mathcal{M}:\substack{g_m(\xi_t)=1,\ n_{m,t}<B_m,\\ \mathcal{A}_{m,t}\neq\varnothing}\right\}.
\label{eq:active_refinement_modes}
\end{equation}
At most one available mode is selected; write $\mu_t\in\mathcal{M}_t$ when one is selected and $\mu_t=\varnothing$ otherwise. The refinement stage returns
\begin{equation}
a_t^{\mathrm{alt}}
=
\begin{cases}
\widetilde a_t^{\mathrm{alt}}, & \mu_t=\varnothing,\\
\displaystyle\operatorname*{arg\,max}_{x\in\mathcal{A}_{\mu_t,t}}h_{\mu_t,t}(x), & \text{otherwise}.
\end{cases}
\label{eq:refined_alternative}
\end{equation}
This case distinction instantiates $\operatorname{Refine}$ in Equation~\ref{eq:care_factorization}.

\paragraph{Exploitation refinement.}
Exploitation refinement searches the program-preferred region for candidates that also receive numerical reference support. Its feasible set takes the form $\mathcal{A}_{\mathrm{exploit},t}=\mathcal{C}_t^{(K)}\cap\mathcal{S}_t^{\mathrm{sup}}$, where $\mathcal{C}_t^{(K)}$ is the top-$K$ subset of $\mathcal{C}_t$ under $z_t$ and $\mathcal{S}_t^{\mathrm{sup}}$ is derived from the support profile $\omega_t$ in $o_t^{\mathrm{ref}}$. The score $h_{\mathrm{exploit},t}$ combines program evidence, reference support, and revealed-trajectory information over this agreement-supported set. This mode concentrates the alternative on candidates where the program and numerical reference provide complementary support.

\paragraph{Exploration refinement.}
Exploration refinement allocates a small candidate budget to coverage of the remaining pool. It selects from an attribute-defined feasible set $\mathcal{A}_{\mathrm{explore},t}$ using a score $h_{\mathrm{explore},t}$ that depends only on pre-selection attributes $\varphi(x)$ and the decision-time context. This mode complements the program's local preference with targeted coverage of the remaining candidate space.

The supplementary material specifies the selector and benchmark-specific mode instantiations. Both refinements only form $a_t^{\mathrm{alt}}$; every refined alternative remains subject to reference-conditioned selection.

\subsection{Reference-Conditioned Selection}
\label{sec:intervention_gate}

The selection stage compares two already constructed candidates; it never generates a third. Let $c_t=(a_t^{\mathrm{ref}},a_t^{\mathrm{alt}},o_t^{\mathrm{ref}},\mathcal{V}_t)$ denote its comparison context. An alternative is eligible only when it is present, unselected, distinct from the reference, and satisfies any active refinement requirements. For an eligible alternative, $\psi_t\in\mathbb{R}$ is a reference-conditioned comparison score that summarizes relative numerical support, score margin, and candidate complementarity; set $\psi_t=-\infty$ otherwise. Selection is
\begin{equation}
\begin{aligned}
\delta_t&=\mathbb{1}\!\left[\psi_t\geq\tau\right],\\
a_t&=
\begin{cases}
a_t^{\mathrm{alt}}, & \delta_t=1,\\
a_t^{\mathrm{ref}}, & \delta_t=0.
\end{cases}
\end{aligned}
\label{eq:gate_decision}
\end{equation}
The case distinction instantiates $\operatorname{Select}$ in Equation~\ref{eq:care_factorization}. It asks whether the alternative meets a reference-conditioned criterion, rather than allowing the scoring program to control the trajectory on its own. The supplementary material specifies the score construction, normalization, eligibility conditions, and threshold used in the reported runs.

\paragraph{Decision record.}
Before observing the outcome, CARE stores $\ell_t=(\pi_t,o_t^{\mathrm{ref}},a_t^{\mathrm{alt}},\delta_t,a_t)$ together with its validation and comparison diagnostics. The record is a by-product of the decision and becomes part of the next-round state only after outcome observation. Its full schema is given in the supplementary material.

\subsection{Procedure for One Round}
\label{sec:per_round_procedure}

Algorithm~\ref{alg:care} gives the temporal order for one replay round.

\begin{algorithm}[tb]
\caption{CARE for one offline replay round}
\label{alg:care}
\begin{algorithmic}[1]
\REQUIRE Decision-time view $\mathcal{V}_t=(\mathcal{H}_t,\mathcal{C}_t,s_t)$ and prior program $\pi_{t-1}$
\STATE $(o_t^{\mathrm{ref}},a_t^{\mathrm{ref}})\leftarrow\mathcal{R}(\mathcal{V}_t)$
\STATE $\pi_t\leftarrow\mathcal{P}(\pi_{t-1},\mathcal{V}_t)$
\IF{$\pi_t\neq\bot$}
  \STATE $a_t^{\mathrm{alt}}\leftarrow\operatorname{Refine}(\pi_t,o_t^{\mathrm{ref}},\mathcal{V}_t)$
\ELSE
  \STATE $a_t^{\mathrm{alt}}\leftarrow\varnothing$
\ENDIF
\STATE Set $\delta_t$ by Equation~\ref{eq:gate_decision}
\STATE $a_t\leftarrow a_t^{\mathrm{alt}}$ if $\delta_t=1$; otherwise $a_t\leftarrow a_t^{\mathrm{ref}}$
\STATE $\ell_t\leftarrow(\pi_t,o_t^{\mathrm{ref}},a_t^{\mathrm{alt}},\delta_t,a_t)$ and diagnostics
\STATE Observe $y_t=f(a_t)$; update $\mathcal{H}_{t+1}$ and $\mathcal{C}_{t+1}$
\STATE $s_{t+1}\leftarrow\mathcal{U}(s_t,\ell_t,(a_t,y_t))$
\end{algorithmic}
\end{algorithm}

%% file: sections/experiments.tex
\section{Experiments}
\label{sec:experiments}

\paragraph{Evaluation questions.}
With five initial and ten adaptive experiments, we test (i) performance against BO and direct-program controls, (ii) the contribution of each controller component, and (iii) sensitivity to the backend and protocol. Every task uses the same fixed-pool replay interface, so the comparison measures budget allocation rather than generation outside the recorded space. All main comparisons use 30 matched seeds.

\paragraph{Chemistry optimization benchmarks.}
We evaluate eight fixed-pool tasks from four benchmark families. \olympusdataset{} is a palladium-catalyzed C--C coupling benchmark with 5{,}670 conditions over ligand, residence time, temperature, and catalyst loading; its objective is emulated yield~\citep{olympus2021,sin2025parallel}. Ligand identity is represented by seven binary variables, one active per condition. \chemlexdataset{} contains 11{,}088 post-deduplication measurements over carboxylic acid, amine, reagent mixture, and solvent; its objective is conversion~\citep{Zhong_2025,chemlexzenodo2025}. The five \bhsuite{} tasks each contain 792 measured yields over aryl halide, additive, ligand, and base~\citep{ahneman2018reaction,hickman2023olympusenhanced}; \directarylation{} contains 1{,}728 yields over ligand, base, solvent, temperature, and concentration~\citep{shields2021edbo,hickman2023olympusenhanced}.

\paragraph{Metrics and statistics.}
We report normalized regret@15, normalized BSF-AUC@10, and Top-1\% Success@15 (Section~\ref{sec:problem}). The latter counts campaigns that evaluate a top $\lceil0.01N\rceil$ condition among the five initial and ten selected experiments. Table~\ref{tab:main_results} reports mean $\pm$ sample standard deviation or successes among 30 campaigns. On \olympus{} and \chemlex{}, paired bootstrap intervals and exact Wilcoxon tests compare CARE with the reference and direct-program controls, with Holm correction over 12 continuous-metric tests. Pool-wide normalization is applied only after replay.

\paragraph{Controls.}
The non-LLM reference policy isolates the reference path. The \emph{evolving scorer (direct)} updates from revealed outcomes but selects its top-ranked condition without a reference; the \emph{fixed scorer (direct)} additionally removes program evolution. Component controls remove planner adaptation or one refinement mode while retaining the rest of the controller. In particular, the planner control removes only CARE's trajectory-derived retain-or-revise override; the round-by-round LLM planner and all selection modules remain active. Table~\ref{tab:ablations} reports trajectory-level changes after mapping each task to $\widetilde y=100(y-y_{\min})/(y_{\max}-y_{\min})$.

\paragraph{Implementation.}
CARE and all LLM controls use GPT-5.4 with matched inference settings, prompts, candidate descriptors, and history access. We froze all implementations before the sweeps. Each baseline selects from its own revealed history and remaining pool under the shared seeds, budget, evaluator, and information constraints. The finite-pool baselines instantiate qLogEI, TPE, Gryffin, EDBO, and LMABO mechanisms~\citep{ament2023logei,botorch2019,tpe2011,gryffin2021,shields2021edbo,ngo2026lmabo}; implementation and cost details are in the supplement.

\begin{table*}[t]
\centering
\caption{Results on eight reaction-optimization tasks over 30 matched seeds. Continuous entries are mean $\pm$ sample standard deviation in percentage points; success entries are campaign counts. Dashes denote unrun method--task pairs.}
\label{tab:main_results}
\footnotesize
\setlength{\tabcolsep}{2.5pt}
\renewcommand{\arraystretch}{1.0}
\newcommand{\meanstd}[2]{\ensuremath{#1\!{\scriptstyle\pm #2}}}
\begin{tabular}{@{}>{\raggedright\arraybackslash}p{0.25\textwidth}*{8}{>{\centering\arraybackslash}p{0.08125\textwidth}}@{}}
\toprule
\multicolumn{9}{@{}l}{\textbf{(a) Normalized regret@15 (\%) $\downarrow$}} \\
\midrule
Method & \shortstack{Suzuki\\(i)} & ChemLex & B--H A & B--H B & B--H C & B--H D & B--H E & \shortstack{Direct\\Arylation} \\
\midrule
\textbf{\method{}} & \meanstd{\mathbf{0.8}}{1.6} & \meanstd{\mathbf{7.9}}{10.1} & \meanstd{\mathbf{1.2}}{1.9} & \meanstd{\mathbf{2.5}}{4.0} & \meanstd{\mathbf{2.0}}{1.7} & \meanstd{\mathbf{1.3}}{1.8} & \meanstd{\mathbf{2.6}}{2.2} & \meanstd{\mathbf{8.3}}{6.4} \\
Non-LLM reference policy & \meanstd{10.3}{11.0} & \meanstd{16.1}{18.1} & \meanstd{8.7}{5.7} & \meanstd{10.2}{9.9} & \meanstd{8.4}{6.2} & \meanstd{4.4}{5.1} & \meanstd{7.5}{5.8} & \meanstd{20.6}{22.6} \\
qLogEI-style & \meanstd{2.8}{3.8} & \meanstd{12.8}{17.8} & \meanstd{8.3}{7.6} & \meanstd{21.0}{9.4} & \meanstd{8.6}{6.3} & \meanstd{6.9}{5.3} & \meanstd{7.2}{6.8} & \meanstd{20.9}{19.6} \\
TPE adaptation & \meanstd{13.0}{9.8} & \meanstd{14.9}{10.2} & \meanstd{12.9}{8.4} & \meanstd{17.7}{10.2} & \meanstd{12.0}{6.0} & \meanstd{10.0}{8.2} & \meanstd{13.5}{8.8} & \meanstd{18.7}{14.2} \\
Gryffin-style & \meanstd{17.5}{13.5} & \meanstd{10.2}{11.8} & \meanstd{10.8}{6.7} & \meanstd{16.5}{10.1} & \meanstd{8.2}{5.1} & \meanstd{5.9}{5.3} & \meanstd{7.7}{5.3} & \meanstd{14.5}{12.5} \\
EDBO-style & \meanstd{17.8}{12.2} & \meanstd{11.1}{7.1} & \meanstd{12.0}{9.0} & \meanstd{19.3}{9.7} & \meanstd{9.8}{5.8} & \meanstd{6.2}{6.4} & \meanstd{10.7}{7.6} & \meanstd{19.8}{16.4} \\
LMABO adaptation & \meanstd{15.4}{11.9} & \meanstd{17.3}{19.2} & \meanstd{13.5}{7.6} & \meanstd{12.0}{9.9} & \meanstd{8.2}{5.3} & \meanstd{7.3}{5.3} & \meanstd{10.4}{6.9} & \meanstd{18.8}{16.2} \\
\midrule
\multicolumn{9}{@{}l}{\textbf{(b) Normalized BSF-AUC@10 (\%) $\uparrow$}} \\
\midrule
Method & \shortstack{Suzuki\\(i)} & ChemLex & B--H A & B--H B & B--H C & B--H D & B--H E & \shortstack{Direct\\Arylation} \\
\midrule
\textbf{\method{}} & \meanstd{\mathbf{94.6}}{3.3} & \meanstd{\mathbf{81.6}}{13.6} & \meanstd{\mathbf{95.6}}{3.1} & \meanstd{\mathbf{92.3}}{5.4} & \meanstd{\mathbf{95.9}}{2.3} & \meanstd{\mathbf{96.1}}{2.3} & \meanstd{\mathbf{95.4}}{2.8} & \meanstd{\mathbf{85.3}}{7.9} \\
Non-LLM reference policy & \meanstd{82.3}{12.7} & \meanstd{77.2}{18.3} & \meanstd{85.9}{6.5} & \meanstd{83.3}{10.7} & \meanstd{87.3}{7.8} & \meanstd{92.7}{5.5} & \meanstd{88.4}{6.4} & \meanstd{68.4}{21.0} \\
qLogEI-style & \meanstd{85.4}{9.4} & \meanstd{76.1}{19.2} & \meanstd{87.2}{7.8} & \meanstd{75.4}{9.8} & \meanstd{88.1}{6.9} & \meanstd{91.0}{6.1} & \meanstd{88.9}{8.8} & \meanstd{67.1}{22.4} \\
TPE adaptation & \meanstd{76.5}{10.2} & \meanstd{78.6}{14.5} & \meanstd{81.7}{8.8} & \meanstd{77.2}{10.8} & \meanstd{86.0}{6.4} & \meanstd{88.3}{8.8} & \meanstd{83.1}{10.6} & \meanstd{69.0}{17.9} \\
Gryffin-style & \meanstd{75.0}{13.3} & \meanstd{77.2}{18.8} & \meanstd{85.2}{7.0} & \meanstd{78.7}{10.3} & \meanstd{88.3}{6.1} & \meanstd{91.6}{5.7} & \meanstd{88.1}{6.3} & \meanstd{71.6}{15.3} \\
EDBO-style & \meanstd{73.6}{13.2} & \meanstd{80.5}{14.6} & \meanstd{83.6}{9.4} & \meanstd{77.2}{11.5} & \meanstd{87.3}{6.2} & \meanstd{91.2}{5.9} & \meanstd{85.5}{8.4} & \meanstd{68.7}{20.1} \\
LMABO adaptation & \meanstd{74.1}{13.2} & \meanstd{72.4}{22.2} & \meanstd{82.7}{6.8} & \meanstd{83.6}{9.0} & \meanstd{86.9}{6.5} & \meanstd{89.6}{6.3} & \meanstd{86.6}{7.9} & \meanstd{70.1}{17.1} \\
\midrule
\multicolumn{9}{@{}l}{\textbf{(c) Top-1\% Success@15 $\uparrow$}} \\
\midrule
Method & \shortstack{Suzuki\\(i)} & ChemLex & B--H A & B--H B & B--H C & B--H D & B--H E & \shortstack{Direct\\Arylation} \\
\midrule
\textbf{\method{}} & \textbf{30/30} & \textbf{8/30} & \textbf{27/30} & \textbf{25/30} & \textbf{26/30} & \textbf{23/30} & \textbf{25/30} & \textbf{18/30} \\
Non-LLM reference policy & 18/30 & 5/30 & 13/30 & 16/30 & 7/30 & 14/30 & 14/30 & 12/30 \\
qLogEI-style & 29/30 & 7/30 & 8/30 & 2/30 & 5/30 & 6/30 & 17/30 & 11/30 \\
TPE adaptation & 13/30 & 0/30 & 3/30 & 5/30 & 3/30 & 5/30 & 2/30 & 8/30 \\
Gryffin-style & 15/30 & 7/30 & 4/30 & 8/30 & 5/30 & 9/30 & 14/30 & 11/30 \\
EDBO-style & 13/30 & 0/30 & 8/30 & 3/30 & 6/30 & 13/30 & 9/30 & 8/30 \\
LMABO adaptation & 10/30 & 3/30 & 4/30 & 12/30 & 5/30 & 4/30 & 5/30 & 9/30 \\
\bottomrule
\end{tabular}
\end{table*}

\paragraph{Overall performance.}
CARE ranks first on all three measures for every task in Table~\ref{tab:main_results}. On \bhsuite{} and \directarylation{}, the strongest non-CARE method by BSF-AUC varies across tasks. Relative to that task-specific comparator, CARE lowers normalized regret by $3.1$--$9.5$ points and raises normalized BSF-AUC by $3.4$--$13.6$ points. It also records the highest Top-1\% Success@15 on each of the six tasks.

Relative to the reference policy on \olympus{}/\chemlex{}, CARE lowers normalized regret by $9.5/8.2$ points and raises normalized BSF-AUC by $12.3/4.4$ points. All four paired comparisons remain significant after Holm correction; bootstrap intervals and tests against both direct-program controls are in the supplement.

For HTE campaigns, final regret measures end-of-budget loss, whereas BSF-AUC rewards finding a strong incumbent earlier; together they distinguish final quality from budget efficiency. This distinction matters under fixed laboratory throughput.

\begin{table}[t]
\centering
\caption{Component controls: increase in regret / decrease in BSF-AUC relative to CARE, in percentage points.}
\label{tab:ablations}
\small
\setlength{\tabcolsep}{2.5pt}
\begin{tabular}{@{}p{0.43\columnwidth}ccc@{}}
\toprule
Control & B--H A & \shortstack{Direct\\Arylation} & \chemlex{} \\
\midrule
Evolving scorer (direct) & 4.2 / 6.2 & 6.5 / 6.4 & 11.0 / 10.4 \\
Fixed scorer (direct) & 6.2 / 7.8 & 13.5 / 12.1 & 10.7 / 5.4 \\
No planner adaptation & 5.1 / 6.0 & 3.9 / 6.0 & 4.7 / 2.9 \\
No exploitation refinement & 1.9 / 3.6 & 8.6 / 8.4 & 6.9 / 3.1 \\
No exploration refinement & 6.9 / 8.0 & 9.8 / 8.9 & 3.5 / 2.4 \\
\midrule
Non-LLM capacity & 5.1 / 6.3 & 7.0 / 5.5 & 10.2 / 6.4 \\
\bottomrule
\end{tabular}
\end{table}

\paragraph{Component controls.}
Every control increases regret and decreases BSF-AUC on all three tasks (Table~\ref{tab:ablations}). The fixed direct scorer further degrades the evolving direct scorer on the two newly evaluated HTE tasks. Exploration refinement contributes most on B--H A and \directarylation{}, whereas exploitation refinement gives the largest component-level regret gain on \chemlex{}. Each control is rerun from initialization, so entries compare complete trajectories.

The gate selects an alternative in 52/300 \olympus{} and 36/300 \chemlex{} decisions, supporting selective intervention rather than direct program control of the trajectory; otherwise, it preserves the reference choice. Most decisions therefore retain the numerical reference trajectory.

\paragraph{A concrete decision trace.}
In \chemlex{} seed 11, round 4, the reference proposed C1 while exploitation refinement formed C2. With aggregate reference rank 2 and support from four rankers, CARE selected C2; its outcome raised the incumbent from 91.71\% to 93.42\%. The supplement gives the conditions and another trace.

\paragraph{Additional checks.}
Across tested initializations and budgets on \olympus{} and \chemlex{}, CARE retains the best final incumbent and BSF-AUC. Replacing GPT-5.4 with Gemini-2.5-Flash changes these measures by only $0.3/1.1$ and $1.3/1.5$ points, respectively. Full trajectories, diagnostics, dispersion, and costs appear in the supplement. Figure~\ref{fig:regret_trajectories} shows when the differences arise: CARE separates early on B--H A and progressively pulls ahead on \chemlex{} as the adaptive budget is used.

\begin{figure}[!b]
\centering
\includegraphics[width=0.98\columnwidth]{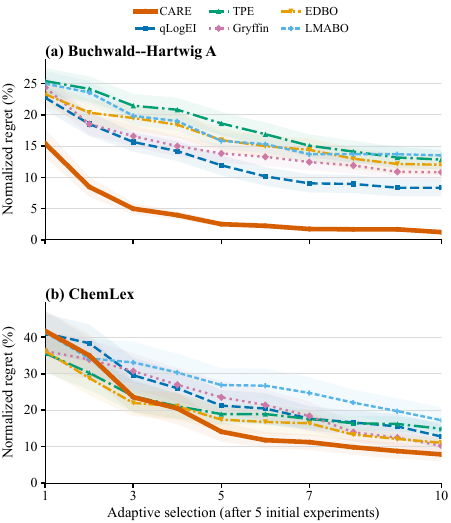}
\caption{Normalized-regret trajectories for CARE and BO baselines on B--H A and \chemlex{} (30 seeds; $\pm1$ s.e. bands).}
\label{fig:regret_trajectories}
\end{figure}

\FloatBarrier

%% file: sections/conclusion.tex
\section{Conclusion}
\label{sec:conclusion}

CARE makes sequential experiment selection a reference-conditioned controller that gates revisable LLM scoring programs against a non-LLM reference candidate. Across the eight evaluated reaction-optimization tasks, it achieves the lowest normalized regret and the highest normalized BSF-AUC and Top-1\% Success@15 among the evaluated methods. Controls show that program evolution is most effective as a selectively gated complement to, rather than a replacement for, the reference policy.

More broadly, CARE retains a numerical optimization anchor while using language models for targeted, evidence-dependent adaptation. The controller is auditable at the decision level: the reference, refinement, and gate each leave a trace that can be inspected before the next experiment.

%% file: sections/limitations.tex
\section*{Limitations}
\label{sec:limitations}

The evaluation uses offline replay on eight finite chemistry tasks. It does not model reagent availability, safety, cycle time, or outcomes outside enumerated pools. It further isolates algorithmic selection from execution: assay noise, failed reactions, and schedule- or inventory-dependent feasibility can reduce the effective budget. CARE is also evaluated only on fixed candidate pools, so its utility for dynamically expanding design spaces remains untested. Prospective use requires a task-specific candidate set, a calibrated reference policy and gate, and laboratory checks on program rankings. Program validation guarantees a well-formed interface, not chemical validity or a replacement for laboratory judgment; robustness to these disruptions requires prospective closed-loop campaigns under realistic laboratory constraints with expert oversight.

%% file: sections/appendix.tex
\appendix

\setlength{\floatsep}{6pt plus 2pt minus 2pt}
\setlength{\textfloatsep}{7pt plus 2pt minus 2pt}
\setlength{\intextsep}{6pt plus 2pt minus 2pt}
\setlength{\dblfloatsep}{7pt plus 2pt minus 2pt}
\setlength{\dbltextfloatsep}{7pt plus 2pt minus 2pt}
\setcounter{topnumber}{4}
\setcounter{bottomnumber}{2}
\setcounter{totalnumber}{6}
\setcounter{dbltopnumber}{3}
\renewcommand{\topfraction}{0.96}
\renewcommand{\bottomfraction}{0.80}
\renewcommand{\textfraction}{0.04}
\renewcommand{\floatpagefraction}{0.75}
\renewcommand{\dbltopfraction}{0.96}
\renewcommand{\dblfloatpagefraction}{0.75}
\makeatletter
\setlength{\@fptop}{0pt}
\setlength{\@fpsep}{8pt plus 2pt minus 2pt}
\setlength{\@fpbot}{0pt plus 1fil}
\setlength{\@dblfptop}{0pt}
\setlength{\@dblfpsep}{8pt plus 2pt minus 2pt}
\setlength{\@dblfpbot}{0pt plus 1fil}
\makeatother

\section{Supplementary Material Overview}
\label{sec:appendix_results}

\begin{figure*}[t]
\centering
\includegraphics[width=0.96\textwidth]{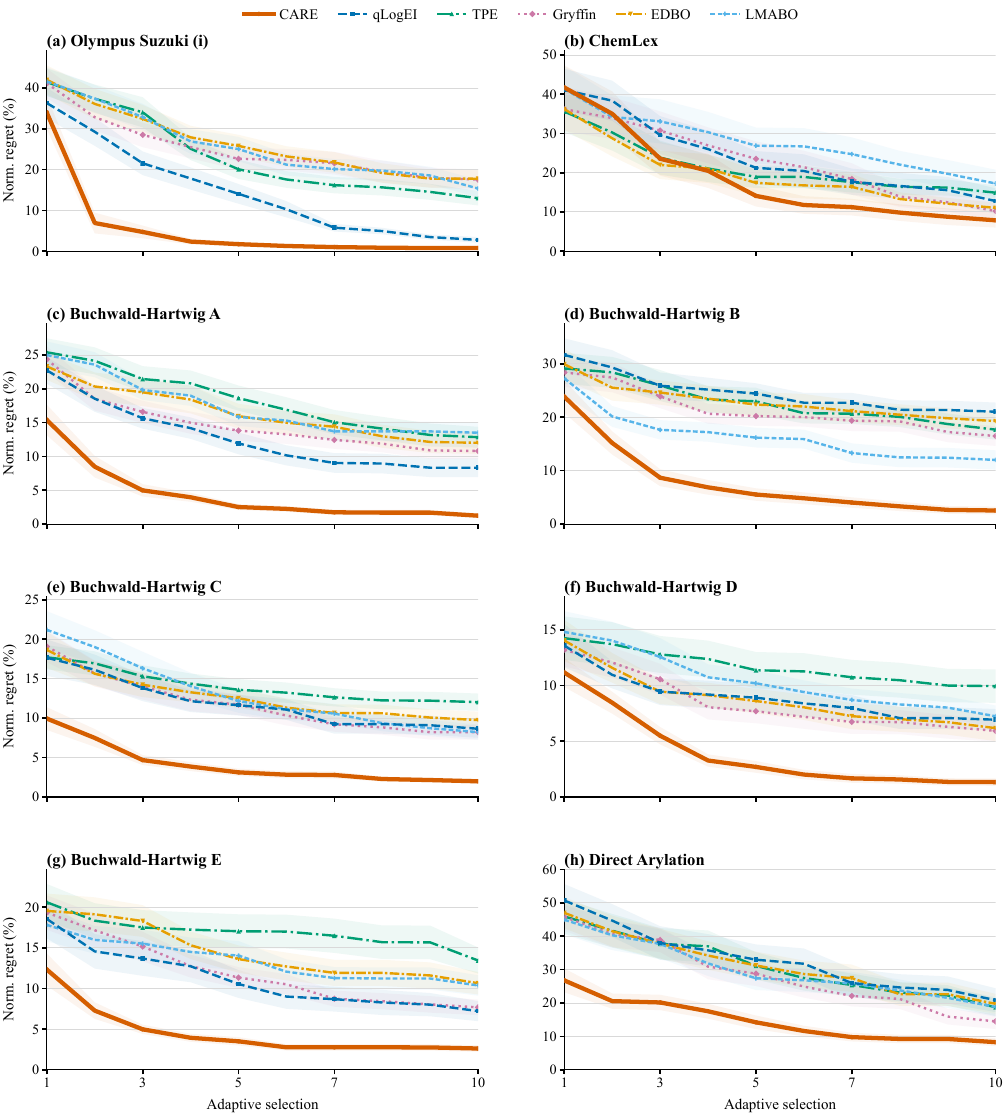}
\caption{Normalized-regret trajectories over all eight reaction-optimization tasks. Curves show means over 30 matched seeds after each of ten adaptive selections following five shared initial observations; shaded bands show $\pm1$ standard error. Lower is better. Every method follows its own revealed history and remaining candidate pool, and the terminal values reproduce the corresponding entries in the main result table.}
\label{fig:supp_all_trajectories}
\end{figure*}

The supplement connects each main-paper claim to its supporting evidence, then provides the configurations and artifacts needed to reproduce the evaluation. Figure~\ref{fig:supplement_roadmap} gives the evidence map; Figure~\ref{fig:supp_all_trajectories} expands the trajectory comparison to all eight tasks. We use the main-paper terminology throughout; monospaced legacy names appear only when reproducing runtime configurations, source files, or prompts and are mapped in Table~\ref{tab:artifact_term_mapping}.

\begin{center}
\setlength{\fboxsep}{3pt}
\footnotesize
\fbox{\parbox{0.90\linewidth}{\textbf{Evidence.} Performance: Figure~\ref{fig:supp_all_trajectories} and Section~\ref{sec:appendix_full_results}. Mechanism and robustness: Section~\ref{sec:appendix_diagnostics}.}}
\par\vspace{3pt}
\fbox{\parbox{0.90\linewidth}{\textbf{Audit and reproduction.} Configurations, seeds, costs, and datasets: Sections~\ref{sec:appendix_repro}--\ref{sec:appendix_protocol}. Decision traces and prompts: Sections~\ref{sec:appendix_cases} and \ref{sec:appendix_prompts}.}}
\captionof{figure}{Supplement roadmap from the paper's claims to the corresponding evidence and reproducibility material.}
\label{fig:supplement_roadmap}
\end{center}
\FloatBarrier

\section{Additional Experiments and Controller Diagnostics}
\label{sec:appendix_diagnostics}
%\label{sec:appendix_diagnostics}

Table~\ref{tab:intervention_diagnostics} summarizes how often the full controller generates and selects alternatives on the two tasks with complete decision diagnostics.

\begin{table}[tbp]
\centering
\caption{Controller activity over 30 matched seeds. An eligible alternative has passed program validation and the checks required for gate evaluation; program versions count distinct validated versions activated within a replay.}
\label{tab:intervention_diagnostics}
\small
\setlength{\tabcolsep}{3pt}
\begin{tabular}{@{}>{\raggedright\arraybackslash}p{0.52\linewidth}>{\centering\arraybackslash}p{0.18\linewidth}>{\centering\arraybackslash}p{0.18\linewidth}@{}}
\toprule
Diagnostic & \olympus{} & \chemlex{} \\
\midrule
Planner calls          & 300 & 300 \\
Rounds with an eligible program alternative & 197 & 225 \\
Selected alternatives  & 52  & 36 \\
Selected alternatives yielding a new best & 21  & 9 \\
Mean distinct active program versions per replay & 6.5 & 7.0 \\
\bottomrule
\end{tabular}
\end{table}

Both exploitation and exploration refinements were active in these runs. The gate selected an alternative in 52 of 300 \olympus{} decisions and 36 of 300 \chemlex{} decisions. On \olympus{}, 34 selected alternatives came from exploitation refinement and 18 from exploration refinement.

\subsection{Complete Component Results}
\label{sec:appendix_complete_components}

Table~\ref{tab:complete_component_results} reports the absolute seed-level results underlying the component deltas in the main paper. The additional no-intervention-gate row changes only the final selector: it retains the reference policy, program proposal, and both refinement modes, but accepts every eligible alternative. This differs from the evolving scorer (direct), which selects from the program ranking without a reference-conditioned comparison. 

Throughout, a ``No X'' row removes component X; a degradation therefore supports the contribution of the removed component. Thus, direct program selection and fixed scoring logic weaken performance, while removing planner adaptation or either refinement mode also weakens the complete controller.

\begin{table*}[t]
\centering
\caption{Complete component and direct-program controls. Except for the marked \chemlex{} capacity cells, entries are mean $\pm$ sample standard deviation over 30 matched seeds and are percentages of the complete-pool outcome range. The capacity control is not a CARE component deletion; its \chemlex{} uncertainty was retained in the available form, mean [95\% paired-bootstrap interval]. Lower normalized regret and higher normalized BSF-AUC are better.}
\label{tab:complete_component_results}
\scriptsize
\setlength{\tabcolsep}{2.0pt}
\renewcommand{\arraystretch}{1.05}
\begin{tabular}{@{}>{\raggedright\arraybackslash}p{0.22\textwidth}*{6}{>{\centering\arraybackslash}p{0.105\textwidth}}@{}}
\toprule
& \multicolumn{2}{c}{B--H A} & \multicolumn{2}{c}{Direct Arylation} & \multicolumn{2}{c}{\chemlex{}} \\
\cmidrule(lr){2-3}\cmidrule(lr){4-5}\cmidrule(lr){6-7}
Method & Regret $\downarrow$ & BSF-AUC $\uparrow$ & Regret $\downarrow$ & BSF-AUC $\uparrow$ & Regret $\downarrow$ & BSF-AUC $\uparrow$ \\
\midrule
\textbf{\method{}} & \textbf{1.24$\pm$1.93} & \textbf{95.61$\pm$3.07} & \textbf{8.28$\pm$6.35} & \textbf{85.26$\pm$7.87} & \textbf{7.87$\pm$10.11} & \textbf{81.57$\pm$13.55} \\
Evolving scorer (direct) & 5.40$\pm$4.32 & 89.46$\pm$5.68 & 14.76$\pm$9.24 & 78.82$\pm$10.92 & 18.84$\pm$19.35 & 71.20$\pm$22.21 \\
Fixed scorer (direct) & 7.44$\pm$4.17 & 87.78$\pm$6.78 & 21.80$\pm$8.70 & 73.12$\pm$12.04 & 18.57$\pm$21.76 & 76.15$\pm$21.37 \\
No planner adaptation & 6.32$\pm$5.15 & 89.62$\pm$5.90 & 12.16$\pm$8.11 & 79.26$\pm$10.03 & 12.61$\pm$12.51 & 78.68$\pm$16.71 \\
No intervention gate & 4.12$\pm$3.93 & 91.61$\pm$5.28 & 16.11$\pm$9.71 & 78.96$\pm$9.87 & 12.43$\pm$12.69 & 79.30$\pm$15.36 \\
No exploitation refinement & 3.18$\pm$2.26 & 92.02$\pm$5.02 & 16.91$\pm$10.86 & 76.82$\pm$12.94 & 14.72$\pm$17.22 & 78.43$\pm$17.85 \\
No exploration refinement & 8.11$\pm$9.90 & 87.58$\pm$11.26 & 18.06$\pm$9.79 & 76.38$\pm$10.31 & 11.41$\pm$11.38 & 79.18$\pm$15.92 \\
\midrule
Non-LLM capacity & 6.35$\pm$4.65 & 89.35$\pm$5.44 & 15.26$\pm$7.76 & 79.80$\pm$9.57 & \shortstack{18.1\\{[12.5, 24.0]}} & \shortstack{75.2\\{[68.4, 81.5]}} \\
\bottomrule
\end{tabular}
\end{table*}

\subsection{Paired Evidence for Primary Comparisons}
\label{sec:appendix_paired_controls}

The paired mean difference favors \method{} in all 12 benchmark, metric, and control combinations (Table~\ref{tab:paired_controls}). We report matched bootstrap intervals, win/tie/loss counts, and two-sided exact Wilcoxon signed-rank tests for the non-LLM reference policy, evolving scorer (direct), and fixed scorer (direct).

\begin{table*}[tbp]
\centering
\caption{Paired comparisons with the primary controls. Deltas are \method{} minus the control over the same 30 seeds, in percentage points of the complete-pool outcome range. CIs are percentile paired bootstrap intervals with 200{,}000 resamples. For normalized regret, W/T/L counts seeds on which CARE has lower, equal, or higher regret; for normalized BSF-AUC it counts higher, equal, or lower values. $p_{\mathrm H}$ is the Holm-adjusted $p$-value from a two-sided exact signed-rank test across the 12 benchmark, metric, and control combinations.}
\label{tab:paired_controls}
\small
\setlength{\tabcolsep}{3.0pt}
{\renewcommand{\arraystretch}{1.08}
\begin{tabular}{@{}>{\raggedright\arraybackslash}p{0.10\textwidth}>{\raggedright\arraybackslash}p{0.20\textwidth}>{\raggedright\arraybackslash}p{0.29\textwidth}>{\raggedright\arraybackslash}p{0.29\textwidth}@{}}
\toprule
Benchmark & Control & Norm. regret $\Delta$ [95\% CI]; W/T/L; $p_{\mathrm H}$ & Norm. BSF-AUC $\Delta$ [95\% CI]; W/T/L; $p_{\mathrm H}$ \\
\midrule
\olympus{} & Non-LLM reference policy & $-9.533$ [$-13.779$, $-5.561$]; 18/8/4; .000247 & +12.272 [8.280, 16.405]; 22/4/4; $3.27{\times}10^{-5}$ \\
& Evolving scorer (direct) & $-22.067$ [$-28.324$, $-16.275$]; 29/1/0; $3.35{\times}10^{-8}$ & +25.715 [20.001, 31.831]; 30/0/0; $2.24{\times}10^{-8}$ \\
& Fixed scorer (direct) & $-28.230$ [$-34.864$, $-22.030$]; 30/0/0; $2.24{\times}10^{-8}$ & +28.624 [22.752, 34.750]; 30/0/0; $2.24{\times}10^{-8}$ \\
\midrule
\chemlex{} & Non-LLM reference policy & $-8.225$ [$-14.735$, $-2.648$]; 10/18/2; .0205 & +4.412 [1.076, 8.332]; 10/16/4; .0491 \\
& Evolving scorer (direct) & $-10.975$ [$-18.774$, $-4.182$]; 20/5/5; .00805 & +10.364 [5.130, 16.025]; 25/1/4; .000411 \\
& Fixed scorer (direct) & $-10.698$ [$-18.536$, $-3.747$]; 18/4/8; .0183 & +5.418 [$-0.628$, 12.128]; 17/2/11; .1256 \\
\bottomrule
\end{tabular}
}
\end{table*}

\subsection{Backend and Gate Threshold Checks}
\label{sec:appendix_backend_margin}

Replacing GPT-5.4 with Gemini-2.5-Flash changes the final incumbent/BSF-AUC by $0.3/1.1$ points on \olympus{} and $1.3/1.5$ points on \chemlex{}, respectively (Table~\ref{tab:backend_robustness}). This second-backend run uses the same 30 matched seeds, 10 rounds, controller, and sampling temperature.

\begin{table}[tbp]
\centering
\caption{Secondary LLM backend run. Values are means over 30 matched seeds on each benchmark's original percentage scale: emulated yield for \olympus{} and reported conversion for \chemlex{}. All rows use 300 planner calls. ``Alt./best'' counts selected alternatives and the subset that yielded a new best after selection.}
\label{tab:backend_robustness}
\small
\setlength{\tabcolsep}{2.2pt}
{\renewcommand{\arraystretch}{1.03}
\begin{tabular}{@{}lcccc@{}}
\toprule
Backend & Final $\uparrow$ & AUC $\uparrow$ & Regret $\downarrow$ & Alt./best \\
\midrule
\multicolumn{5}{@{}l}{\olympus{}} \\
GPT-5.4 & 88.5 & 84.4 & 0.7 & 52 / 21 \\
Gemini-2.5-Flash & 88.2 & 83.3 & 1.0 & 42 / 20 \\
\midrule
\multicolumn{5}{@{}l}{\chemlex{}} \\
GPT-5.4 & 92.1 & 81.6 & 7.9 & 36 / 9 \\
Gemini-2.5-Flash & 90.8 & 80.1 & 9.2 & 35 / 7 \\
\bottomrule
\end{tabular}
}
\end{table}

Across $\tau=-0.1$ to $+0.1$, final best and AUC each vary by at most 0.3 points (Table~\ref{tab:gate_margin_sensitivity}). Raising $\tau$ to $+0.2$ leaves final best unchanged relative to $+0.1$ and lowers AUC by 3.5 points.

\begin{table}[tbp]
\centering
\caption{\olympus{} selection-threshold sensitivity over 10 matched seeds. An eligible alternative is selected when $\psi_t\geq\tau$. Final, AUC, and regret are in percentage points of emulated yield. The main configuration uses $\tau=0.0$. This independent rerun resamples scoring programs, so its zero-threshold row is distinct from the main-sweep estimate.}
\label{tab:gate_margin_sensitivity}
\small
\setlength{\tabcolsep}{2.4pt}
{\renewcommand{\arraystretch}{1.05}
\begin{tabular}{@{}lcccc@{}}
\toprule
Threshold $\tau$ & Final $\uparrow$ & AUC $\uparrow$ & Regret $\downarrow$ & Alternatives \\
\midrule
$-0.1$ & 86.8 & 83.5 & 2.5 & 18 \\
$0.0$ & 86.6 & 83.6 & 2.6 & 18 \\
$+0.1$ & 86.5 & 83.8 & 2.7 & 17 \\
$+0.2$ & 86.5 & 80.3 & 2.7 & 19 \\
\bottomrule
\end{tabular}
}
\end{table}

\subsection{Replay Protocol Variations}
\label{sec:appendix_protocol_robustness}

\method{} attains the best terminal incumbent and BSF-AUC in every tested initialization and budget setting on the two protocol-sweep tasks, \olympus{} and \chemlex{}. The four settings vary the initial observations and subsequent selections as (5,10), (3,10), (5,20), and (10,10). Tables~\ref{tab:protocol_robustness_olympus} and~\ref{tab:protocol_robustness_chemlex} give all method means.

\begin{table*}[t]
\centering
\caption{Full replay protocol check on \olympusdataset{}. Final, AUC, and regret are means over 10 matched seeds per protocol in percentage points of emulated yield. The variations change the number of initial observations and selections while keeping the controller fixed.}
\label{tab:protocol_robustness_olympus}
\small
\setlength{\tabcolsep}{2pt}
{\renewcommand{\arraystretch}{1.02}
\begin{tabular*}{0.94\textwidth}{@{\extracolsep{\fill}}llccc@{}}
\toprule
Protocol & Method & Final best $\uparrow$ & Best-so-far AUC $\uparrow$ & Regret $\downarrow$ \\
\midrule
\multicolumn{5}{l}{\textbf{Main (5 init, 10 selections)}} \\
 & \textbf{\method{}} & \textbf{88.1} & \textbf{84.7} & \textbf{1.2} \\
 & Non-LLM reference policy & 83.0 & 77.6 & 6.3 \\
 & Generic surrogate BO & 80.9 & 74.7 & 8.3 \\
 & Classical GP-UCB & 79.9 & 74.0 & 9.3 \\
 & Fixed scorer (direct) & 72.2 & 69.4 & 17.0 \\
\midrule
\multicolumn{5}{l}{\textbf{Fewer initial observations (3 init, 10 selections)}} \\
 & \textbf{\method{}} & \textbf{88.4} & \textbf{82.6} & \textbf{0.8} \\
 & Non-LLM reference policy & 81.6 & 75.3 & 7.6 \\
 & Generic surrogate BO & 79.4 & 72.3 & 9.8 \\
 & Classical GP-UCB & 78.5 & 71.8 & 10.8 \\
 & Fixed scorer (direct) & 70.8 & 67.5 & 18.4 \\
\midrule
\multicolumn{5}{l}{\textbf{Longer budget (5 init, 20 selections)}} \\
 & \textbf{\method{}} & \textbf{89.2} & \textbf{86.2} & \textbf{0.0} \\
 & Non-LLM reference policy & 85.8 & 80.7 & 3.4 \\
 & Generic surrogate BO & 83.3 & 77.6 & 6.0 \\
 & Classical GP-UCB & 82.4 & 76.7 & 6.8 \\
 & Fixed scorer (direct) & 75.5 & 72.0 & 13.7 \\
\midrule
\multicolumn{5}{l}{\textbf{More initial observations (10 init, 10 selections)}} \\
 & \textbf{\method{}} & \textbf{88.6} & \textbf{86.3} & \textbf{0.7} \\
 & Non-LLM reference policy & 85.9 & 82.6 & 3.3 \\
 & Generic surrogate BO & 83.8 & 79.2 & 5.5 \\
 & Classical GP-UCB & 83.0 & 78.5 & 6.2 \\
 & Fixed scorer (direct) & 75.1 & 72.7 & 14.1 \\
\bottomrule
\end{tabular*}
}
\end{table*}

\begin{table*}[t]
\centering
\caption{Full replay protocol check on \chemlexdataset{}. Final, AUC, and regret are means over 10 matched seeds per protocol in percentage points of reported conversion. The variations change the number of initial observations and selections while keeping the controller fixed.}
\label{tab:protocol_robustness_chemlex}
\small
\setlength{\tabcolsep}{2pt}
{\renewcommand{\arraystretch}{1.02}
\begin{tabular*}{0.94\textwidth}{@{\extracolsep{\fill}}llccc@{}}
\toprule
Protocol & Method & Final best $\uparrow$ & Best-so-far AUC $\uparrow$ & Regret $\downarrow$ \\
\midrule
\multicolumn{5}{l}{\textbf{Main (5 init, 10 selections)}} \\
 & \textbf{\method{}} & \textbf{96.2} & \textbf{88.6} & \textbf{3.8} \\
 & Non-LLM reference policy & 94.5 & 86.8 & 5.5 \\
 & No intervention gate & 92.4 & 86.8 & 7.6 \\
 & Gryffin-style & 93.4 & 79.1 & 6.6 \\
 & EDBO-style & 90.1 & 79.3 & 9.9 \\
 & Fixed scorer (direct) & 81.7 & 78.9 & 18.3 \\
\midrule
\multicolumn{5}{l}{\textbf{Fewer initial observations (3 init, 10 selections)}} \\
 & \textbf{\method{}} & \textbf{95.3} & \textbf{86.0} & \textbf{4.7} \\
 & Non-LLM reference policy & 92.9 & 83.7 & 7.1 \\
 & No intervention gate & 91.4 & 84.1 & 8.6 \\
 & Gryffin-style & 91.8 & 76.0 & 8.2 \\
 & EDBO-style & 88.7 & 76.6 & 11.3 \\
 & Fixed scorer (direct) & 79.8 & 75.9 & 20.2 \\
\midrule
\multicolumn{5}{l}{\textbf{Longer budget (5 init, 20 selections)}} \\
 & \textbf{\method{}} & \textbf{97.8} & \textbf{91.6} & \textbf{2.2} \\
 & Non-LLM reference policy & 96.2 & 89.9 & 3.8 \\
 & No intervention gate & 95.3 & 90.3 & 4.7 \\
 & Gryffin-style & 94.9 & 82.5 & 5.1 \\
 & EDBO-style & 92.5 & 83.2 & 7.5 \\
 & Fixed scorer (direct) & 84.4 & 81.2 & 15.6 \\
\midrule
\multicolumn{5}{l}{\textbf{More initial observations (10 init, 10 selections)}} \\
 & \textbf{\method{}} & \textbf{96.9} & \textbf{91.2} & \textbf{3.1} \\
 & Non-LLM reference policy & 95.7 & 89.8 & 4.3 \\
 & No intervention gate & 94.0 & 89.3 & 6.0 \\
 & Gryffin-style & 94.6 & 83.7 & 5.4 \\
 & EDBO-style & 92.0 & 83.9 & 8.0 \\
 & Fixed scorer (direct) & 84.0 & 81.4 & 16.0 \\
\bottomrule
\end{tabular*}
}
\end{table*}

\subsection{Decision Trace, Validation, and Cost}
\label{sec:appendix_repro}

\paragraph{Decision trace.}
\label{sec:appendix_audit_ledger}
\method{} records the artifacts that determine every selection before its outcome is observed (Table~\ref{tab:audit_ledger}). In the main-paper notation, this record instantiates the transition from $\mathcal{V}_t$ to $\ell_t$ and, after observation, to $s_{t+1}$. A scoring program must compile, pass restricted execution, assign a finite score to every remaining candidate, and preserve its scores under candidate reordering. A validated program supplies the initial proposal; exploitation or exploration refinement may form an alternative; the gate then compares that alternative with the reference.

\paragraph{Information boundary and LLM cost.}
The information boundary permits revealed outcomes, candidate attributes available before selection, and the recorded state of the optimizer. Planner prompts contain trajectory, candidate, controller-state, and previous-gate summaries; synthesis prompts request executable \texttt{rank\_candidates} code. Once validated, that code scores the full remaining pool without another LLM call. Complete calibrated token accounting is available for \olympus{} and \chemlex{} (Table~\ref{tab:api_cost}); the other six tasks did not retain the same token-accounting fields, so we report the observed two-task average rather than extrapolating it. Appendix~\ref{sec:appendix_prompts} reproduces the canonical prompts.

\begin{table}[tbp]
\centering
\caption{CARE API usage for the two fully instrumented tasks. Total-token values are calibrated estimates obtained from recorded prompt and response sizes, including visible output and reasoning tokens. The combined row sums 60 campaigns; its per-campaign columns are averages over those campaigns.}
\label{tab:api_cost}
\small
\setlength{\tabcolsep}{2.2pt}
\renewcommand{\arraystretch}{1.04}
\begin{tabular}{@{}lrrrrr@{}}
\toprule
Task & Plan & Synth. & Calls & Calls/run & Tokens/run \\
\midrule
\olympus{} & 300 & 197 & 497 & 16.6 & 124.4k \\
\chemlex{} & 300 & 225 & 525 & 17.5 & 136.8k \\
\midrule
Combined / mean & 600 & 422 & 1{,}022 & 17.0 & 130.6k \\
\bottomrule
\end{tabular}
\end{table}

The corresponding estimated totals are 3.73M tokens for \olympus{}, 4.10M for \chemlex{}, and 7.84M across both. The calibration uses the reported GPT-5.4 Responses configuration with medium reasoning and low response verbosity. Monetary totals are omitted because they depend on the pricing schedule rather than the replay protocol.

\method{} has no local language-model training stage: local computation consists of updating the reference models and executing validated ranking programs over the finite candidate pool. The archived runs do not contain comparable wall-clock, CPU-hour, or accelerator-hour counters, so we do not infer them from file timestamps. The reproducible cost account is therefore the fixed experiment budget together with the measured call counts and calibrated token workload above.

The supplementary material documents the replay configurations, matched seed schedules, result summaries, decision traces, scoring-program examples, prompt templates, the \chemlex{} duplicate-handling procedure, and the software environment.

\begin{table}[tbp]
\centering
\caption{Contents of the decision trace. ``Pre'' records are written before the outcome is observed, while ``Post'' denotes the subsequent outcome update. Together these records reconstruct the selected candidate from the information available at decision time.}
\label{tab:audit_ledger}
\small
\setlength{\tabcolsep}{2.0pt}
{\renewcommand{\arraystretch}{1.05}
\begin{tabular}{@{}>{\raggedright\arraybackslash}p{0.25\linewidth}>{\raggedright\arraybackslash}p{0.12\linewidth}>{\raggedright\arraybackslash}p{0.54\linewidth}@{}}
\toprule
Artifact & Time & Stored fields and role \\
\midrule
Prompt payload & Pre & Summaries of the trajectory, candidates, persistent state, and previous gate report; defines the interface visible to the LLM. \\
Generated scoring program & Pre & Source hash, identifier, output schema, and validation status; records the generated rule. \\
Program validation & Pre & Compilation result, restricted execution, complete finite scoring, and invariance to candidate order; determines program eligibility. \\
Reference candidate & Pre & Candidate from the non-LLM reference policy and its support metadata; fixes the default selection. \\
Alternative candidate & Pre & When a validated program is available, the candidate formed by retaining its initial proposal or applying exploitation or exploration refinement. \\
Gate record & Pre & When an alternative is available, its features, support counts, comparison with the reference, gate score, and decision. \\
Selection record & Pre & Reference, available alternative, selected candidate, and the reason for the selection. \\
Outcome update & Post & Revealed outcome and best-so-far update; affects only later decisions and evaluation. \\
\bottomrule
\end{tabular}
}
\end{table}

The trace makes each decision replayable from its recorded input summaries, program and validation result, available candidates, gate computation, and final selection. Empty alternative and gate records indicate deterministic fallback to the reference policy when validation produces no eligible program.

\subsection{Controller Implementation Summary}
\label{sec:appendix_care_impl}

\noindent This section instantiates the main-paper interfaces $\mathcal{R}$, $\mathcal{P}$, $\operatorname{Refine}$, and $\operatorname{Select}$, together with the state transition $\mathcal{U}$. We fixed all numerical settings before the 30-seed sweep and reused them across seeds and rounds. The two fully instrumented controller configurations share the selection-score coefficients and thresholds; each activates the reference rankers and refinement rules supported by its task attributes. Every adaptive update uses only the recorded state available when the decision is made. 

Program evolution modifies only the scoring program; the non-LLM reference policy remains a separate, unchanged comparison path.

Before round $t$, $\bar b_t=\max\{y:(x,y)\in\mathcal H_t\}$ is the best revealed outcome, and $y_{t-1}=f(a_{t-1})$ is the latest revealed outcome when $t>1$. The aggregate reference score defines rank $R_t(x)$ and the top-40 reference shortlist $\mathcal T_t$. In the main-paper notation, the support profile $\omega_t(x)$ records membership in the leading shortlist of each active reference ranker. Each reference ranker also stores its own top-40 shortlist for agreement features and a stricter top-12 shortlist for exploitation eligibility.

For the gate features, let $m_t$ be the maximum of 1, the interquartile range, and the standard deviation of the current aggregate reference scores. If refinement assigns a score $h_t(x)$ to a proposal, its normalized refinement score is $\tanh(\max(0,h_t(x))/m_t)$. The reference score margin compares the alternative and reference under $S_t^{\mathrm{ref}}$ and divides the difference by $m_t$. The combined margin uses the same normalization after the scoring program rank is added to the reference aggregation. Ranker disagreement is $(d_t-1)/(n_t-1)$, where $d_t$ is the number of distinct top choices among the $n_t$ active reference rules; it is zero when fewer than two rules are active. A strongly supported reference is one whose normalized categorical evidence is at least 0.75; its penalty increases with the gap between that evidence and the alternative's categorical evidence.

Novelty is the candidate's Euclidean distance to its nearest revealed condition after the permitted attributes are encoded and standardized; distances are rescaled to $[0,1]$ over the current candidates. The categorical feature score aggregates, across the available categorical fields, shrunk means, best revealed values, support counts, and the frequency of low outcomes for each observed category value. This aggregate is also rescaled to $[0,1]$ over the current candidates. Unseen category values receive a small default contribution rather than an outcome estimate.

\begin{table*}[tbp]
\centering
\caption{Controller implementation for the main replay suite. Adaptive quantities use revealed outcomes, candidate attributes available before selection, and recorded controller state. Normalized ranker agreement compares the number of individual shortlists containing the alternative with the largest count in $\mathcal T_t$. The bounded Olympus exploration feature is $\operatorname{clip}_{[0,1]}(0.45L_3+0.25T+0.25C+0.05(1-|D-0.10|))$.}
\label{tab:care_fixed_config}
\small
\setlength{\tabcolsep}{4.0pt}
{\renewcommand{\arraystretch}{1.05}
\begin{tabular}{@{}>{\raggedright\arraybackslash}p{0.18\textwidth}>{\raggedright\arraybackslash}p{0.76\textwidth}@{}}
\toprule
Component & Reported instantiation \\
\midrule
Alternative eligibility & The scoring program must pass compilation, restricted execution, complete finite scoring, and invariance checks for candidate order. The candidate must remain unselected and differ from the reference. A refinement mode must also pass its budget, trigger, and support checks. \\
Mode selection & At most one refinement forms the alternative in a round. The reported controller evaluates exploration refinement first and then exploitation refinement if no exploration candidate satisfies the refinement checks; otherwise it retains the initial program proposal. \\
Selection score & For an eligible alternative, the reported comparison score is $\psi_t=\boldsymbol{\alpha}^{\top}\mathbf{g}_t-\boldsymbol{\beta}^{\top}\mathbf{q}_t$, where $\mathbf{g}_t$ and $\mathbf{q}_t$ contain positive and penalty features. Select the alternative iff $\psi_t\geq\tau$, before observing the selected outcome. The score is unitless, and the main runs use $\tau=0$, the equality point between the positive and penalty totals. \\
Positive gate features & $0.38$ normalized refinement score; $0.24$ positive reference score margin; $0.18$ positive combined score margin; $0.14$ normalized ranker agreement; $0.10/\sqrt{R_t(a_t^{\mathrm{alt}})}$ aggregate rank bonus; $0.08$ novelty; $0.06$ categorical feature score; and $0.14$ exploration feature score. \\
Penalty gate features & $0.24$ when the alternative lies outside the reference shortlist $\mathcal T_t$; $0.20$ when no individual ranker supports it; $0.10$ when support comes only from the scoring program outside the exploration rule; $0.28$ times disagreement with a strongly supported reference; and $0.08$ times a negative reference score margin. \\
Program evolution & $\mathcal{P}$ is restricted to retention, revision, or regeneration. Its summary uses the last five revealed objectives and last three selection reports. In the first round, it requests a new program and does not apply a latest-outcome rule. Later, it recommends revision after at least two consecutive rounds without improvement or when $y_{t-1}$ trails $\bar b_t$ by at least $\max(15,0.30|\bar b_t|)$; a recent improvement recommends retention. A newly produced program must pass the checks in the alternative-eligibility row; otherwise, the controller retains a prior valid program or returns $\bot$. \\
Validated-program selection & The action above determines whether new source is synthesized. The program used for ranking is then chosen from the new artifact, up to four recent validated artifacts, and the first validated artifact by revealed-only prequential replay. Each of at most five folds runs the program with a prefix of $\mathcal H_t$ as observed data and scores its ranking of a future revealed suffix as $0.70$ times the selected outcome plus $0.25$ times the top-3 mean outcome minus $0.05$ times the rank of the best suffix outcome. Fold outcomes are withheld while ranking, and stored memory and tool state are not reused. A newly revised program is rejected in favor of the prior active program if either comparison has no valid fold or if its score is worse by more than $0.02$ times the revealed outcome range. When the evolution action is retention, the highest-scoring validated program is used, with deterministic ties. This revealed-history selector is part of $\mathcal P$ in the main-paper notation. \\
Exploitation refinement & Budget 2 per replay. The reported instantiation starts in round 4, scans the top 8 candidates from the scoring program, and activates when $\bar b_t$ is below 75, the latest outcome is far below $\bar b_t$, or $\bar b_t$ is below 84 after at least two stagnant rounds with ranker disagreement of at least 0.72. A candidate must have aggregate reference rank at most 4 or appear in the top-12 shortlists of at least two reference rankers. Its refinement score $h_t(x)$ must reach 1.2. The ChemLex instantiation may start in round 3 and scans up to 12 candidates. Its model-band branch requires support from both GP and RF rules and from both categorical rules, and it does not displace a reference backed by categorical evidence of at least 0.85. Eligible candidates are ordered by a fixed combination of model ranks, categorical evidence, normalized reference score, novelty, and the trajectory trigger. \\
Exploration refinement & Budget 1 per replay and active in both configurations summarized here. A proposal can be formed from round 2 when $35\leq\bar b_t<78$ and, if $\bar b_t>72$, only after an outcome far below the best. A benchmark-specific exploration score ranks unused conditions. The Olympus instantiation scores L3 conditions as $3.00L_3+1.25T+1.00C+0.18(1-|D-0.10|)$, where $L_3$ is the ligand indicator and $T$, $C$, and $D$ are normalized temperature, catalyst loading, and residence time. The ChemLex instantiation applies a categorical exploration ranking over acid, amine, reagent, and solvent. Every exploration proposal still requires the intervention gate. \\
LLM backend & OpenAI-compatible Responses API, \texttt{gpt-5.4}, reasoning effort \texttt{medium}, temperature 1.0, structured JSON outputs, verbosity \texttt{low}, and max output-token limits of 6000/9000 for \olympus{}/\chemlex{}. \\
\bottomrule
\end{tabular}
}
\end{table*}

\begin{table*}[t]
\centering
\caption{Non-LLM reference policy used to compute $S_t^{\mathrm{ref}}$. Base weights, update rules, and schedules are prespecified; realized weights change only with revealed outcomes.}
\label{tab:nonllm_optimizer_config}
\small
\setlength{\tabcolsep}{3.0pt}
{\renewcommand{\arraystretch}{1.05}
\begin{tabular}{@{}>{\raggedright\arraybackslash}p{0.20\textwidth}>{\raggedright\arraybackslash}p{0.74\textwidth}@{}}
\toprule
Reference policy component & Fixed instantiation and role \\
\midrule
Reference rankers & Fixed heuristic, GP-EI, RF-UCB, categorical shrinkage, and categorical empirical Bayes UCB when their required attributes are available. The scoring program is separate and can change the selection only through an accepted alternative. \\
Rank transform & Each ranker contributes $w_{t,e}/\sqrt{r_{t,e}(x)}$ over its top-40 shortlist. Exploitation checks use the stricter stored top-12 shortlist. Realized weights use revealed observations only and are clamped to $[0.60,1.60]$ before schedule multipliers. \\
Phase schedule & Early phase emphasizes heuristic/prior weights; mid and late phases emphasize surrogate and categorical replay weights. \\
Condition prior $P_t$ & Normalized temperature, catalyst loading, residence time, and ligand identity evidence when those fields exist. \\
Novelty/density & Normalized distance from revealed observations and density among the remaining candidates over ligand identity when available. \\
Categorical statistics $Q_t$ & Shrunk component statistics from revealed observations: means, best observed values, support counts, and penalties for low outcomes. \\
Sparse categorical route & In categorical spaces with many values, categorical evidence receives higher weight and route switching emphasizes model UCB or empirical Bayes evidence from revealed statistics. \\
\bottomrule
\end{tabular}
}
\end{table*}

The six-rule capacity control is separate from the reference policy in Table~\ref{tab:nonllm_optimizer_config}. It combines a fixed heuristic, GP expected improvement, random forest upper confidence bound, categorical shrinkage, EDBO descriptor GP expected improvement, and BayBE/BoFire mixed variable scoring in the same rank aggregation framework. It contains no scoring program, program evolution planner, proposal refinement, or intervention gate.

\subsection{Exact Replay and Runtime Configuration}
\label{sec:appendix_exact_config}

Table~\ref{tab:exact_runtime_config} records the literal settings used by the two fully instrumented CARE runs. The artifact schema predates the paper terminology: \texttt{macro\_frontier\_scout} denotes exploration refinement and \texttt{residual\_scout} denotes exploitation refinement. All settings were fixed before the 30-seed evaluation.

\begin{table*}[t]
\centering
\caption{Exact CARE replay and runtime configuration. A semicolon separates literal configuration values. The task-specific scoring formulas, gate coefficients, and eligibility definitions are given in Table~\ref{tab:care_fixed_config}.}
\label{tab:exact_runtime_config}
\scriptsize
\setlength{\tabcolsep}{3.0pt}
\renewcommand{\arraystretch}{1.06}
\begin{tabular}{@{}>{\raggedright\arraybackslash}p{0.18\textwidth}>{\raggedright\arraybackslash}p{0.38\textwidth}>{\raggedright\arraybackslash}p{0.38\textwidth}@{}}
\toprule
Field & \olympus{} & \chemlex{} \\
\midrule
Dataset and objective & \texttt{minerva\_csv}; maximize \texttt{yield}; 5{,}670 candidates & \texttt{chemlex\_csv}; maximize \texttt{Conversion}; 11{,}088 candidates after cleaning \\
Campaigns and budget & seeds 0--29; 5 initial observations; 10 adaptive selections & seeds 0--29; 5 initial observations; 10 adaptive selections \\
Initial observations & Sort by \texttt{candidate\_id}; sample five IDs without replacement with \texttt{numpy.default\_rng(seed)}; sort the sampled IDs & Same construction after the fixed cleaning and row-shuffle steps \\
Data identity & Raw rows; sequential candidate IDs & Group duplicate conditions by the four reaction fields and average; drop groups with replicate range $\geq50$; deterministic hash IDs; row-shuffle seed 0 \\
LLM interface & Responses API; GPT-5.4; temperature 1.0; medium reasoning; low verbosity & Same \\
Response execution & maximum output 6{,}000 tokens; streaming on; timeout 360\,s & maximum output 9{,}000 tokens; streaming off; timeout 360\,s \\
Parsing and repair & 2 API parse retries; 1 program-repair attempt; retain the last valid active program after parse failure & 2 API parse retries; 2 program-repair attempts; retain the last valid active program after parse failure \\
Gate & calibrated comparison; margin $\tau=0$ & Same \\
Exploration refinement & enabled; budget 1; first eligible round 2; best-value thresholds 35/72/78 & Same, with categorical exploration ranking over acid, amine, reagent, and solvent \\
Exploitation refinement & enabled; budget 2; first eligible round 4; program top-$k=8$; reference rank $\leq4$ or support $\geq2$; minimum refinement score 1.2 & Same general branch; categorical model-band branch from round 3, scan up to 12, categorical anchor guard 0.85 \\
State and fallback & Fresh state per campaign; validated program retention is permitted; deterministic reference fallback when no eligible alternative exists & Same \\
\bottomrule
\end{tabular}
\end{table*}

Every method receives the same five initial IDs for a task and seed. Subsequent tie-breaking sorts equal scores by \texttt{candidate\_id}. The official qLogEI run sets the Torch seed to $10{,}000s+t$ for campaign seed $s$ and round $t$; the other numerical methods likewise receive recorded campaign- and round-specific seeds. LLM generations remain sampled at temperature 1.0, and the exact returned programs, hashes, prompt payloads, and selections are retained in the trace.

\section{Baseline Results and Dispersion}
\label{sec:appendix_full_results}

\method{} has the lowest normalized regret, the highest normalized BSF-AUC, and the highest Top-1\% Success@15 on all eight tasks. Tables~\ref{tab:full_olympus} and~\ref{tab:full_chemlex} report the complete baseline suites for those tasks with bootstrap intervals; Table~\ref{tab:bh_da_dispersion} reports sample standard deviations for the shared strong-baseline panel on \bhsuite{} and \directarylation{}.

\subsection{Baseline Configuration and Parity}
\label{sec:appendix_baseline_parity}

All rows share the same initial observations and candidate set, a budget of 10 selections, the objective scale, seed schedule, evaluator, and deterministic rule for resolving tied scores. Their trajectories then diverge as they select different conditions. Baselines use permitted candidate attributes and outcomes revealed on their own trajectories. Nonadaptive rows use candidate identifiers and strata available at that point. Surrogate, descriptor, categorical, mixed space BO, and LMABO adaptations fit on revealed observations and score all candidates remaining on their own trajectories. The LMABO adaptation preserves acquisition function selection and evaluates the chosen acquisition over those candidates. Controls involving scoring programs share \method{}'s language model, history summary format, candidate attributes, and prompt interface. \method{} variants disable one named component while retaining the other components, configurations, and state update rules. The controlled resource is the experiment budget; Section~\ref{sec:appendix_repro} separately reports the available API resource account and the boundary of the retained compute records. Each adapted baseline uses one fixed benchmark-compatible configuration across all seeds.

The controller controls isolate distinct choices. \emph{No planner adaptation} retains the round-by-round LLM planner but removes CARE's trajectory-derived override of its retain-or-revise choice. \emph{No intervention gate} retains planner adaptation, both refinement modes, and the reference policy, but allows every eligible alternative to bypass gate filtering. The \emph{evolving scorer (direct)} omits CARE's comparison between a reference and an alternative; the evolving scoring program selects the next condition directly. The \emph{fixed scorer (direct)} is generated once and its source code is not revised. It is rerun in every round on the growing revealed history and selects directly, without the reference policy or gate. These rows compare complete controller variants. A method name containing ``adaptation'' retains the named family's central representation, surrogate, or acquisition rule in the finite candidate setting.

\begin{center}
\begin{minipage}{\linewidth}
\centering
\captionof{table}{Shared conditions for the complete baseline suite.}
\label{tab:baseline_parity_summary}
\small
\setlength{\tabcolsep}{2.7pt}
{\renewcommand{\arraystretch}{1.05}
\begin{tabular}{@{}p{0.27\linewidth}p{0.64\linewidth}@{}}
\toprule
Row family & Shared parity constraint \\
\midrule
Random/heuristic & Same seed schedule, candidate identifiers, available strata, and deterministic resolution of tied scores. \\
Surrogate BO & Fit only on revealed observations and score every candidate remaining on that method's trajectory. \\
Chemistry BO & Use only permitted descriptors or categorical fields, never outcomes held by the evaluator. \\
LMABO adaptation & Use the current BO state to select an acquisition function, then score the candidates remaining on that method's trajectory. \\
Scoring program controls & Same prompt interface and language model; differ in program evolution and gate usage. \\
\method{} variants & Disable one named component while retaining the other components, configurations, and state update rules; mean degradations appear in the main-paper component table. \\
\bottomrule
\end{tabular}
}
\end{minipage}
\end{center}

\begin{table*}[t]
\centering
\caption{Frozen configurations for the strong baseline panel in the main table and Figure~\ref{fig:supp_all_trajectories}. All methods refit on their own revealed history, score the full remaining pool, and resolve equal scores by \texttt{candidate\_id}. Fixed kernel parameters mean that no hidden hyperparameter search is performed during replay.}
\label{tab:strong_baseline_exact_config}
\scriptsize
\setlength{\tabcolsep}{3.0pt}
\renewcommand{\arraystretch}{1.06}
\begin{tabular}{@{}>{\raggedright\arraybackslash}p{0.18\textwidth}>{\raggedright\arraybackslash}p{0.30\textwidth}>{\raggedright\arraybackslash}p{0.46\textwidth}@{}}
\toprule
Method / tasks & Representation and model & Acquisition and fixed settings \\
\midrule
qLogEI-style; \olympus{}/\chemlex{} & BoTorch 0.16.1 under Python 3.11.15 and Torch 2.13.0; \texttt{SingleTaskGP} with input normalization for numeric \olympus{}, \texttt{MixedSingleTaskGP} over all categorical dimensions for \chemlex{}; standardized outcomes & $q=1$ \texttt{qLogExpectedImprovement}; exact marginal-likelihood fit with \texttt{maxiter=100}; Torch seed $10{,}000s+t$; score every unrevealed candidate \\
qLogEI-style; B--H/Direct Arylation & Standardized permitted features with 64 hashed dimensions and cap 96 (48/cap 64 in high-cardinality spaces); Gaussian process with constant $\times$ Mat\'ern-$5/2$ plus white noise & Constant and length-scale bounds $[0.05,20]$; white-noise level $10^{-3}$ with bounds $[10^{-6},1]$; $\alpha=10^{-6}$; normalized target; optimizer disabled; log expected improvement \\
TPE adaptation & Permitted numeric and string-derived features; 128 hashed dimensions, cap 160; standardized from revealed observations & Good set is the upper 35\% of revealed utility (0.65 quantile); score $0.82$ good-vs.-bad distance-ratio rank $+0.18$ diversity rank; deterministic $10^{-9}$ jitter \\
Gryffin-style & Permitted categorical/string features; 192 hashed dimensions, cap 224; 192-tree extra-trees ensemble, square-root feature subsampling, bootstrap on & Score $0.48$ categorical evidence $+0.24$ mean $+0.18$ ensemble standard deviation $+0.10$ diversity; before three observations use $0.72$ categorical evidence $+0.28$ diversity \\
EDBO-style & Descriptor/string features; 256 hashed dimensions, cap 256; standardized Mat\'ern-$5/2$ GP with fixed length scale 1, fixed white noise $10^{-4}$, $\alpha=10^{-6}$, normalized target, optimizer disabled & Score $0.62$ EI $+0.18$ posterior mean $+0.12$ categorical evidence $+0.08$ diversity; before three observations use $0.55$ categorical evidence $+0.45$ diversity \\
LMABO adaptation & GPT-5.4 chooses among PI, LogPI, EI, LogEI, UCB, PosMean, PosSTD, TS, qKG, qPES, qMES, and qJES; 128 hashed permitted features, cap 160; fixed Mat\'ern-$5/2$ GP blended 0.70/0.30 with 5-nearest-neighbor estimates & One acquisition-choice call per adaptive round (10 per campaign); the chosen acquisition is evaluated over the remaining pool and blended with nearest-neighbor mean (0.10), diversity (0.05), and categorical evidence (0.03); implementation reference \texttt{giang-n-ngo/lmabo@b1671f5ad} \\
\bottomrule
\end{tabular}
\end{table*}

\paragraph{Summary of complete results.}
Random and heuristic rows calibrate the finite candidate task, BO rows instantiate standard optimizer families, and program controls compare CARE with the evolving scorer (direct) and fixed scorer (direct). Top-1\% Success@15 means that at least one of the five initial or ten adaptive selections belongs to the $\lceil0.01N\rceil$ highest-outcome conditions in the complete benchmark; it is computed only after replay.

\begin{table*}[t]
\centering
\caption{Complete matched replay suite on \olympusdataset{}. Normalized regret and normalized BSF-AUC are percentages of the complete-pool outcome range and report mean [95\% bootstrap CI] over 30 matched seeds. Top-1\% Success@15 reports successful campaigns. Named BO adaptations use the same finite candidate interface.}
\label{tab:full_olympus}
\small
\setlength{\tabcolsep}{3.2pt}
\begin{tabular}{@{}>{\raggedright\arraybackslash}p{0.38\textwidth}ccc@{}}
\toprule
Method & Norm. regret@15 (\%) $\downarrow$ & Norm. BSF-AUC@10 (\%) $\uparrow$ & Top-1\% Success@15 $\uparrow$ \\
\midrule
\textbf{\method{}} & \textbf{0.8 [0.3, 1.4]} & \textbf{94.6 [93.3, 95.6]} & \textbf{30/30} \\
Non-LLM reference policy & 10.3 [6.6, 14.3] & 82.3 [77.7, 86.6] & 18/30 \\
qLogEI-style~\citep{ament2023logei,botorch2019} & 2.8 [1.6, 4.3] & 85.4 [81.9, 88.5] & 29/30 \\
TPE adaptation~\citep{tpe2011} & 13.0 [9.7, 16.5] & 76.5 [72.9, 80.1] & 13/30 \\
LMABO adaptation~\citep{ngo2026lmabo} & 15.4 [11.4, 19.6] & 74.1 [69.4, 78.5] & 10/30 \\
BayBE/BoFire adaptation~\citep{baybe2025,bofire2024} & 15.7 [12.6, 19.4] & 75.8 [71.9, 79.4] & 11/30 \\
Gryffin-style~\citep{gryffin2021} & 17.5 [13.0, 22.5] & 75.0 [70.2, 79.5] & 15/30 \\
EDBO-style~\citep{shields2021edbo} & 17.8 [13.6, 22.1] & 73.6 [69.0, 78.2] & 13/30 \\
GP-EI~\citep{jones1998ego} & 18.5 [13.9, 23.4] & 73.4 [67.8, 78.7] & 13/30 \\
GP-UCB~\citep{srinivas2010gpucb} & 15.5 [10.9, 20.6] & 74.8 [69.4, 79.9] & 13/30 \\
BoTorch adaptation~\citep{botorch2019} & 18.8 [14.1, 23.7] & 73.3 [67.7, 78.6] & 12/30 \\
Generic surrogate BO~\citep{breiman2001randomforests,auer2002finitebandit} & 16.1 [11.9, 20.6] & 71.4 [65.5, 77.1] & 13/30 \\
SMAC adaptation~\citep{smac2011} & 28.7 [21.9, 36.2] & 63.4 [55.9, 70.4] & 5/30 \\
Chemistry descriptor BO~\citep{gryffin2021,shields2021edbo} & 21.6 [17.4, 26.4] & 73.2 [68.3, 77.9] & 5/30 \\
Categorical empirical Bayes UCB~\citep{robbins1956empiricalbayes,auer2002finitebandit} & 28.4 [22.4, 35.0] & 66.3 [60.1, 72.1] & 6/30 \\
Fixed non-LLM heuristic & 22.6 [16.3, 29.3] & 74.5 [68.2, 80.5] & 11/30 \\
Random & 31.5 [26.5, 36.7] & 63.5 [57.5, 69.4] & 1/30 \\
Stratified random & 32.4 [25.8, 39.1] & 64.4 [58.3, 70.3] & 4/30 \\
Evolving scorer (direct) & 22.8 [17.1, 29.1] & 68.9 [62.3, 75.0] & 10/30 \\
Fixed scorer (direct) & 29.0 [22.9, 35.6] & 65.9 [59.4, 72.2] & 5/30 \\
\bottomrule
\end{tabular}
\end{table*}

\begin{table*}[t]
\centering
\caption{Complete matched replay suite on \chemlexdataset{}. Normalized regret and normalized BSF-AUC are percentages of the complete-pool outcome range and report mean [95\% bootstrap CI] over 30 matched seeds. Top-1\% Success@15 reports successful campaigns. Named BO adaptations use the same finite candidate interface.}
\label{tab:full_chemlex}
\small
\setlength{\tabcolsep}{3.2pt}
\begin{tabular}{@{}>{\raggedright\arraybackslash}p{0.38\textwidth}ccc@{}}
\toprule
Method & Norm. regret@15 (\%) $\downarrow$ & Norm. BSF-AUC@10 (\%) $\uparrow$ & Top-1\% Success@15 $\uparrow$ \\
\midrule
\textbf{\method{}} & \textbf{7.9 [4.6, 11.7]} & \textbf{81.6 [76.7, 86.3]} & \textbf{8/30} \\
Non-LLM reference policy & 16.1 [10.1, 22.7] & 77.2 [70.5, 83.4] & 5/30 \\
qLogEI-style~\citep{ament2023logei,botorch2019} & 12.8 [7.2, 19.5] & 76.1 [69.0, 82.4] & 7/30 \\
TPE adaptation~\citep{tpe2011} & 14.9 [11.5, 18.7] & 78.6 [73.2, 83.4] & 0/30 \\
BayBE/BoFire adaptation~\citep{baybe2025,bofire2024} & 16.9 [11.1, 23.5] & 74.7 [67.7, 81.2] & 5/30 \\
Gryffin-style~\citep{gryffin2021} & 10.2 [6.4, 14.6] & 77.2 [70.4, 83.6] & 7/30 \\
EDBO-style~\citep{shields2021edbo} & 11.1 [8.7, 13.7] & 80.5 [74.9, 85.2] & 0/30 \\
LMABO adaptation~\citep{ngo2026lmabo} & 17.3 [10.8, 24.2] & 72.4 [64.5, 79.8] & 3/30 \\
GP-EI~\citep{jones1998ego} & 13.8 [9.3, 18.7] & 75.0 [67.3, 82.0] & 4/30 \\
GP-UCB~\citep{srinivas2010gpucb} & 16.9 [11.2, 23.6] & 75.5 [67.5, 82.6] & 1/30 \\
BoTorch adaptation~\citep{botorch2019} & 18.7 [11.3, 27.5] & 72.8 [63.4, 81.1] & 5/30 \\
Generic surrogate BO~\citep{breiman2001randomforests,auer2002finitebandit} & 23.2 [12.5, 35.5] & 70.4 [58.6, 81.0] & 5/30 \\
SMAC adaptation~\citep{smac2011} & 13.7 [9.4, 18.8] & 71.0 [61.5, 79.7] & 4/30 \\
Chemistry descriptor BO~\citep{gryffin2021,shields2021edbo} & 17.6 [10.8, 25.9] & 73.5 [65.3, 80.8] & 4/30 \\
Categorical empirical Bayes UCB~\citep{robbins1956empiricalbayes,auer2002finitebandit} & 24.6 [15.9, 34.3] & 68.2 [57.3, 78.1] & 2/30 \\
Fixed non-LLM heuristic & 10.5 [9.8, 11.0] & 73.8 [68.7, 78.9] & 0/30 \\
Random & 18.1 [12.8, 24.3] & 73.7 [66.6, 80.1] & 1/30 \\
Stratified random & 21.4 [14.3, 29.5] & 71.1 [63.3, 78.1] & 3/30 \\
Evolving scorer (direct) & 18.8 [12.6, 26.2] & 71.2 [63.1, 78.7] & 3/30 \\
Fixed scorer (direct) & 18.6 [11.7, 26.8] & 76.1 [68.1, 83.1] & 3/30 \\
\bottomrule
\end{tabular}
\end{table*}

\begin{table*}[t]
\centering
\caption{Matched replay on \bhsuite{} and \directarylation{}. Continuous entries are mean $\pm$ sample standard deviation over 30 matched seeds and percentages of the complete-pool outcome range. Top-1\% Success@15 reports successful campaigns. The non-LLM reference row uses the same reference-policy implementation as the full CARE controller on these tasks.}
\label{tab:bh_da_dispersion}
\footnotesize
\setlength{\tabcolsep}{2.5pt}
\renewcommand{\arraystretch}{1.25}
\begin{tabular}{@{}>{\raggedright\arraybackslash}p{0.22\textwidth}*{6}{c}@{}}
\toprule
\multicolumn{7}{@{}l}{\textbf{(a) Normalized regret@15 (\%) $\downarrow$}} \\
\midrule
Method & B--H A & B--H B & B--H C & B--H D & B--H E & \shortstack{Direct\\Arylation} \\
\midrule
\textbf{\method{}} & \textbf{1.24$\pm$1.93} & \textbf{2.54$\pm$4.01} & \textbf{1.98$\pm$1.69} & \textbf{1.33$\pm$1.81} & \textbf{2.62$\pm$2.19} & \textbf{8.28$\pm$6.35} \\
Non-LLM reference policy & 8.73$\pm$5.69 & 10.24$\pm$9.90 & 8.41$\pm$6.19 & 4.42$\pm$5.10 & 7.52$\pm$5.76 & 20.56$\pm$22.61 \\
qLogEI-style~\citep{ament2023logei,botorch2019} & 8.33$\pm$7.59 & 21.02$\pm$9.43 & 8.64$\pm$6.28 & 6.94$\pm$5.25 & 7.23$\pm$6.77 & 20.86$\pm$19.62 \\
TPE adaptation~\citep{tpe2011} & 12.85$\pm$8.44 & 17.67$\pm$10.22 & 12.02$\pm$5.98 & 9.95$\pm$8.21 & 13.45$\pm$8.79 & 18.72$\pm$14.19 \\
Gryffin-style~\citep{gryffin2021} & 10.81$\pm$6.67 & 16.47$\pm$10.11 & 8.20$\pm$5.07 & 5.94$\pm$5.32 & 7.69$\pm$5.25 & 14.50$\pm$12.53 \\
EDBO-style~\citep{shields2021edbo} & 12.02$\pm$8.98 & 19.27$\pm$9.67 & 9.75$\pm$5.84 & 6.18$\pm$6.37 & 10.66$\pm$7.57 & 19.77$\pm$16.36 \\
LMABO adaptation~\citep{ngo2026lmabo} & 13.51$\pm$7.62 & 12.00$\pm$9.89 & 8.22$\pm$5.31 & 7.25$\pm$5.28 & 10.39$\pm$6.92 & 18.84$\pm$16.20 \\
\midrule
\multicolumn{7}{@{}l}{\textbf{(b) Normalized BSF-AUC@10 (\%) $\uparrow$}} \\
\midrule
Method & B--H A & B--H B & B--H C & B--H D & B--H E & \shortstack{Direct\\Arylation} \\
\midrule
\textbf{\method{}} & \textbf{95.61$\pm$3.07} & \textbf{92.25$\pm$5.40} & \textbf{95.90$\pm$2.25} & \textbf{96.10$\pm$2.28} & \textbf{95.43$\pm$2.78} & \textbf{85.26$\pm$7.87} \\
Non-LLM reference policy & 85.92$\pm$6.50 & 83.34$\pm$10.68 & 87.26$\pm$7.75 & 92.69$\pm$5.46 & 88.37$\pm$6.42 & 68.43$\pm$21.04 \\
qLogEI-style & 87.21$\pm$7.84 & 75.44$\pm$9.76 & 88.14$\pm$6.89 & 91.04$\pm$6.05 & 88.86$\pm$8.80 & 67.07$\pm$22.38 \\
TPE adaptation & 81.74$\pm$8.81 & 77.24$\pm$10.83 & 85.98$\pm$6.40 & 88.30$\pm$8.80 & 83.08$\pm$10.62 & 68.96$\pm$17.91 \\
Gryffin-style & 85.24$\pm$7.04 & 78.72$\pm$10.25 & 88.25$\pm$6.09 & 91.55$\pm$5.73 & 88.07$\pm$6.28 & 71.63$\pm$15.30 \\
EDBO-style & 83.60$\pm$9.44 & 77.15$\pm$11.49 & 87.33$\pm$6.22 & 91.20$\pm$5.85 & 85.52$\pm$8.37 & 68.71$\pm$20.12 \\
LMABO adaptation & 82.68$\pm$6.75 & 83.55$\pm$8.98 & 86.93$\pm$6.49 & 89.59$\pm$6.28 & 86.58$\pm$7.89 & 70.09$\pm$17.07 \\
\midrule
\multicolumn{7}{@{}l}{\textbf{(c) Top-1\% Success@15 $\uparrow$}} \\
\midrule
Method & B--H A & B--H B & B--H C & B--H D & B--H E & \shortstack{Direct\\Arylation} \\
\midrule
\textbf{\method{}} & \textbf{27/30} & \textbf{25/30} & \textbf{26/30} & \textbf{23/30} & \textbf{25/30} & \textbf{18/30} \\
Non-LLM reference policy & 13/30 & 16/30 & 7/30 & 14/30 & 14/30 & 12/30 \\
qLogEI-style & 8/30 & 2/30 & 5/30 & 6/30 & 17/30 & 11/30 \\
TPE adaptation & 3/30 & 5/30 & 3/30 & 5/30 & 2/30 & 8/30 \\
Gryffin-style & 4/30 & 8/30 & 5/30 & 9/30 & 14/30 & 11/30 \\
EDBO-style & 8/30 & 3/30 & 6/30 & 13/30 & 9/30 & 8/30 \\
LMABO adaptation & 4/30 & 12/30 & 5/30 & 4/30 & 5/30 & 9/30 \\
\bottomrule
\end{tabular}
\par\vspace{7pt}
\begin{minipage}{0.92\textwidth}
\footnotesize
\textit{Reading the matched results.}
The Buchwald--Hartwig tasks differ only in the underlying reaction table, yet their strongest non-CARE comparator changes across A--E; this makes performance across the suite more informative than a single favorable instance. CARE retains low regret and high BSF-AUC throughout the suite, while Direct Arylation remains harder for every method. Its advantage is also visible in campaign-level Top-1\% success rather than only in the means. Because every row replays the same 30 initial candidate sets within a task, the corresponding seed-wise comparisons are paired; the displayed standard deviations describe campaign variability, not uncertainty of the mean.
\end{minipage}
\end{table*}

\FloatBarrier
\twocolumn[{
\begin{center}
\includegraphics[width=0.74\textwidth]{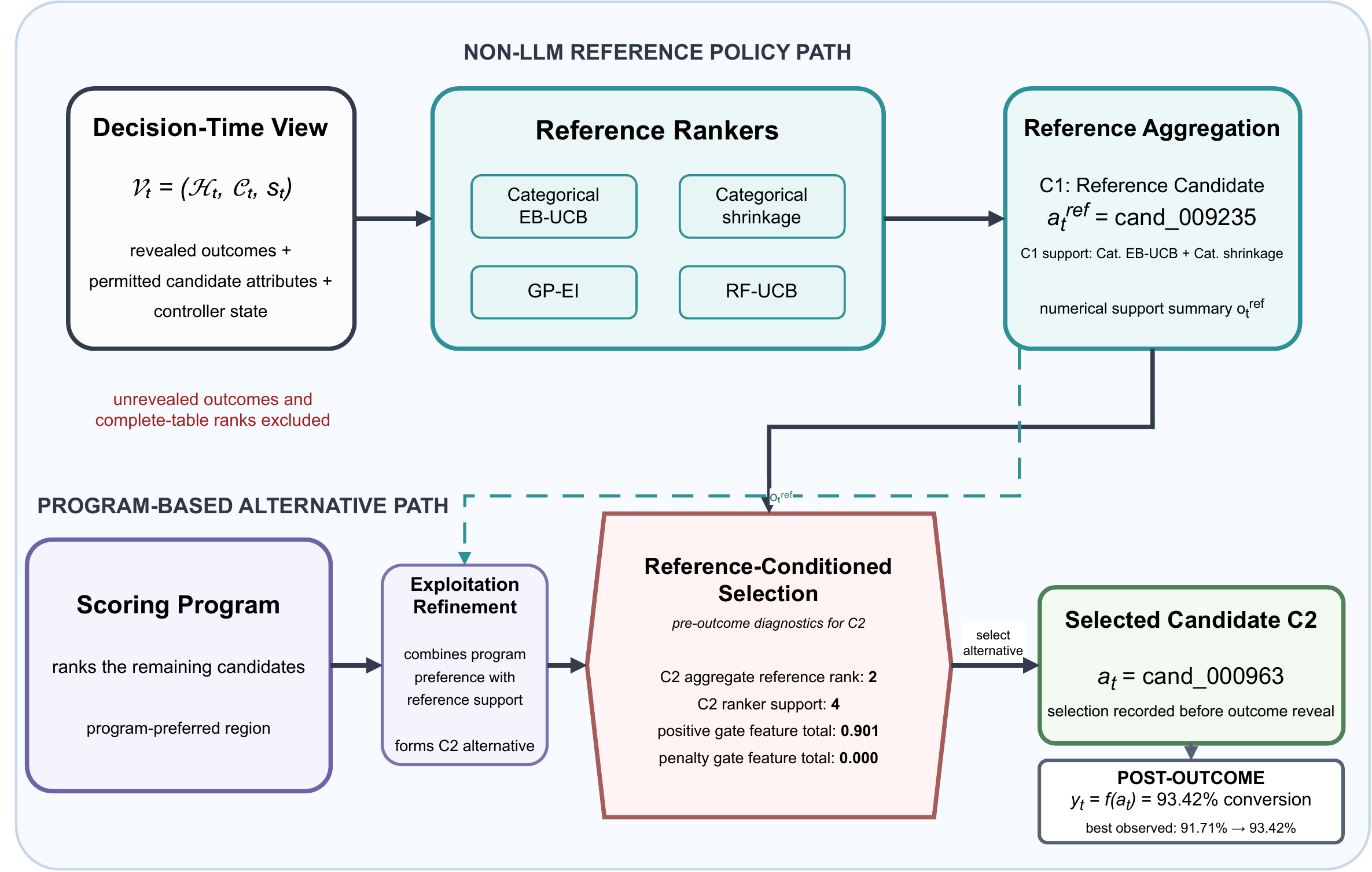}
\captionof{figure}{\chemlex{} seed 11, round 4 decision trace. The non-LLM reference policy proposes C1, and exploitation refinement forms C2 as the program-based alternative. The intervention gate records the candidates, support signals, and gate decision before revealing C2's 93.42\% conversion. Table~\ref{tab:chemlex_case_conditions} expands the candidate IDs.}
\label{fig:chemlex_case}
\end{center}
\vspace{2pt}
}]

\section{Representative Replay Traces}
\label{sec:appendix_cases}

Each trace separates the decision record written before selection from the outcome revealed afterward. Outcomes of unselected reference candidates and full-table ranks, which use unrevealed outcomes, are added only for retrospective comparison.

\paragraph{Reference and selected alternative on \chemlex{}.}
\chemlex{} seed 11, round 4 illustrates a decision in a large categorical space. The non-LLM reference policy proposed C1, supported by categorical empirical Bayes UCB and categorical shrinkage rankers. Exploitation refinement formed C2 from candidates favored by the scoring program and submitted it as the alternative. Before its outcome was revealed, the decision trace stored aggregate reference rank 2, support from four distinct rankers, positive gate feature total 0.901, and penalty gate feature total 0.000. The gate selected C2. Its reported conversion was then revealed as 93.42\%, raising the best observed conversion from 91.71\% to 93.42\%. Figure~\ref{fig:chemlex_case} shows the corresponding artifact trace, and Table~\ref{tab:chemlex_case_conditions} expands the candidate IDs into reaction conditions.

\begin{table*}[t]
\centering
\caption{Concrete \chemlex{} conditions from the replay trace. C1, C2, and C3 are local aliases for conditions visible at decision time. All three share the same reagent mixture and solvent, while the alternatives change the carboxylic acid and amine pair. Outcomes are retrospective: C2 and C3 were revealed after selection, whereas C1 was not selected.}
\label{tab:chemlex_case_conditions}
\small
\setlength{\tabcolsep}{2.2pt}
{\renewcommand{\arraystretch}{1.00}
\begin{tabular}{@{}>{\raggedright\arraybackslash}p{0.16\textwidth}>{\raggedright\arraybackslash}p{0.11\textwidth}>{\raggedright\arraybackslash}p{0.31\textwidth}>{\raggedright\arraybackslash}p{0.25\textwidth}>{\raggedleft\arraybackslash}p{0.09\textwidth}@{}}
\toprule
Alias and role & Display ID & Acid & Amine & Benchmark conversion (\%) \\
\midrule
C1: reference candidate & \path|cand_009235| & \smiles{COc1ccc(C=CC(=O)Nc2ccccc2C(=O)O)cc1OC} & \smiles{Nc1ccc(Br)c2cccnc12} & 0.00 \\
C2: round-4 alternative & \path|cand_000963| & \smiles{CC(C)[C@H](NC(=O)OCC1c2ccccc2-c2ccccc21)C(=O)O} & \smiles{Cc1cc(N2CCNCC2)n(-c2ccccc2)n1} & 93.42 \\
C3: round-6 alternative & \path|cand_001842| & \smiles{O=C(N[C@H](COCc1ccccc1)C(=O)O)OCC1c2ccccc2-c2ccccc21} & \smiles{Cc1cc(N2CCNCC2)n(-c2ccccc2)n1} & 94.74 \\
\midrule
\multicolumn{5}{@{}>{\raggedright\arraybackslash}p{0.96\textwidth}@{}}{\textit{Shared permitted attributes:} reagent mixture \smiles{CCN(C(C)C)C(C)C.CN(C)C(On1nnc2cccnc21)=[N+](C)C.F[P-](F)(F)(F)(F)F}; solvent \smiles{CN(C)C=O}.} \\
\bottomrule
\end{tabular}
}
\end{table*}

The same trajectory later produced a second exploitation alternative. At round 6, the reference policy again proposed C1, while exploitation refinement proposed C3. The decision trace stored support from four rankers, aggregate reference rank 4, positive gate feature total 1.023, and penalty gate feature total 0.000 before the outcome was known. The gate selected C3, whose reported conversion was then revealed as 94.74\%. Together, these two rounds show how exploitation refinement used program rankings and reference support within one replay.

\paragraph{Exploration refinement on \olympus{}.}
In seed 3, round 4, exploration refinement submits M2 and the gate selects it. M2 uses L3, a longer residence time, and a higher temperature than the L0 reference M1; its revealed yield is 89.23\%, compared with the unselected reference's retrospective value of 64.55\% (Table~\ref{tab:olympus_case_conditions}).

\begin{table}[tbp]
\centering
\caption{\olympus{} conditions for seed 3, round 4. The source table stores ligand identity in seven one-hot indicator columns, shown here as L0/L3. Outcomes are retrospective: M2 was revealed after selection, whereas M1 was not selected in this round.}
\label{tab:olympus_case_conditions}
\small
\setlength{\tabcolsep}{3.0pt}
{\renewcommand{\arraystretch}{1.06}
\begin{tabular}{@{}>{\raggedright\arraybackslash}p{0.28\linewidth}>{\raggedright\arraybackslash}p{0.28\linewidth}>{\raggedright\arraybackslash}p{0.28\linewidth}@{}}
\toprule
Field & M1: reference candidate & M2: alternative candidate \\
\midrule
Display ID & \path|cand_000072| & \path|cand_002592| \\
Ligand identity & L0 & L3 \\
Residence time & 60.0 & 120.0 \\
Temperature & 100.0 & 110.0 \\
Catalyst loading & 2.515 & 2.515 \\
Retrospective emulated yield (\%) & 64.55 & 89.23 \\
\bottomrule
\end{tabular}
}
\end{table}

\paragraph{A non-improving accepted intervention.}
\olympus{} seed 20, round 4 provides a counterexample to the successful cases above. After two rounds without incumbent improvement, exploitation refinement proposed N2 (\path|cand_004940|) instead of reference N1 (\path|cand_002511|). Before either outcome was available to the selector, N2 had aggregate reference rank 2, support from two non-LLM rankers in addition to the scoring program, gate score 0.862, and zero recorded penalty; the gate therefore accepted it. N2's subsequently revealed yield was 78.79\%, below the current 81.79\% incumbent, so the intervention produced no improvement. For retrospective comparison only, N1's benchmark yield is 87.00\%. This trace shows that the gate is an evidence-based selector rather than an outcome oracle and makes a false-positive intervention directly auditable.

\begin{table}[tbp]
\centering
\caption{Representative selected alternatives, including one non-improving intervention. Rank, ranker support, and gate evidence are recorded before outcome reveal; conversion or emulated yield is shown afterward.}
\label{tab:case_audit_records}
\small
\setlength{\tabcolsep}{3.0pt}
{\renewcommand{\arraystretch}{1.08}
\begin{tabular}{@{}>{\raggedright\arraybackslash}p{0.46\linewidth}>{\raggedright\arraybackslash}p{0.46\linewidth}@{}}
\toprule
Decision & Gate features before selection and outcome afterward \\
\midrule
\chemlex{}, seed 11/round 4: C2 replaces C1 (Table~\ref{tab:chemlex_case_conditions}) & rank 2; support 4; positive 0.901; penalty 0.000; conversion 93.42\% \\
\chemlex{}, seed 11/round 6: C3 replaces C1 (Table~\ref{tab:chemlex_case_conditions}) & rank 4; support 4; positive 1.023; penalty 0.000; conversion 94.74\% \\
\olympus{}, seed 3/round 4: M2 replaces M1 (Table~\ref{tab:olympus_case_conditions}) & rank 1; support 2; positive 0.662; penalty 0.044; emulated yield 89.23\% \\
\olympus{}, seed 20/round 4: N2 replaces N1 & rank 2; support 2; gate score 0.862; penalty 0.000; emulated yield 78.79\%; no new best \\
\bottomrule
\end{tabular}
}
\end{table}

\subsection{Outcome-Guided Program Evolution in One Replay}
\label{sec:appendix_scoring_program_cases}

The adaptive budget is not a batch size. Each replay starts from five observations and then executes ten $q=1$ rounds: one condition is selected, its outcome is revealed, and only then can the program state change for the next decision. The regenerate/revise/retain action determines whether new source is synthesized; the revealed-only validated-program selector described in Table~\ref{tab:care_fixed_config} then chooses the program that supplies the current ranking. Consequently, retention means that no new source is generated in that round, although an earlier validated program can become active again when its prequential score is highest.

\begin{figure*}[tbp]
\centering
\includegraphics[width=0.91\textwidth]{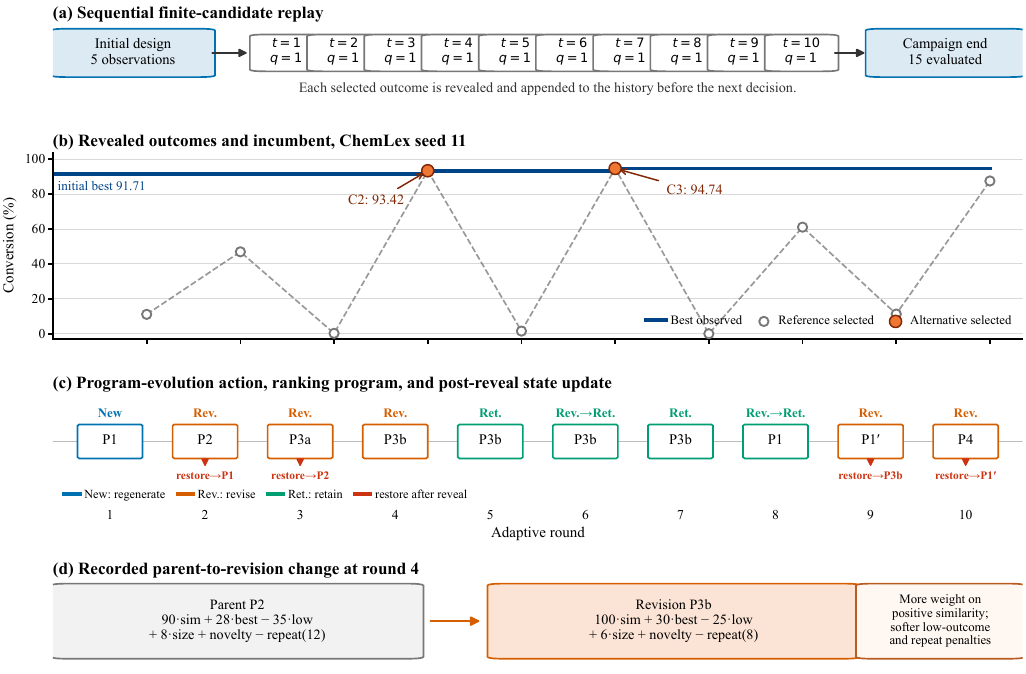}
\caption{Sequential protocol and outcome-guided program evolution for the audited \chemlex{} seed-11 replay. (a) $B=10$ denotes ten one-at-a-time adaptive selections after five initial observations, not ten-condition batches; every outcome is revealed before the next round. (b) Gray markers are selected conversions, the blue step curve is the best revealed conversion, and orange markers are the gate-selected alternatives C2 and C3. (c) Each box is the scoring program that supplied the round's ranking. New/Rev./Ret. denote regeneration, revision, and retention; Rev.$\rightarrow$Ret. means that the trajectory-derived evolution override retained a program instead of applying a suggested revision. Restore marks a post-reveal state update that reactivates a prior validated program. (d) The compact semantic diff reproduces the principal coefficient changes from P2 to its round-4 revision P3b; ``sim,'' ``best,'' and ``low'' denote outcome-weighted positive similarity, similarity to the best revealed pair, and similarity to revealed nonpositive pairs. Input parsing, stable sorting, and diagnostics are omitted.}
\label{fig:program_evolution_timeline}
\end{figure*}

Figure~\ref{fig:program_evolution_timeline} separates three events that are easy to conflate: source generation before selection, the validated scoring program used to rank the current pool, and the state update after outcome reveal. Table~\ref{tab:chemlex_program_lineage} gives the corresponding round-level provenance. In rounds 6 and 8, the trajectory-derived evolution override applies retention instead of a suggested revision, so no new source is synthesized. At round 8, the revealed-only selector reactivates P1 from the validated-program portfolio. Rounds 2, 3, 9, and 10 use a newly revised program for the current decision, then restore a prior validated program after the selected outcome underperforms the pre-round incumbent; this post-reveal update cannot alter the already-completed decision.

This replay exhibits the intended division of labor. Only two of the ten adaptive selections use an alternative, and both establish a new incumbent: C2 raises conversion from 91.71\% to 93.42\%, and C3 raises it again to 94.74\%. The same P3b scoring program supplies both rankings, while reference-conditioned selection retains the reference candidate in the other rounds. The final incumbent therefore improves by 3.03 percentage points while program revisions and post-reveal restoration remain explicit in the recorded state.

\begin{table*}[tbp]
\centering
\caption{Complete scoring-program lineage for \chemlex{} seed 11. $|\mathcal H_t|$ is the number of revealed observations available before selection. ``Suggested$\rightarrow$applied'' reports the evolution action before and after the trajectory-derived evolution override; identical actions are shown once. The program-used column is the artifact that ranked the current pool, whereas the post-reveal column is the state retained for subsequent rounds. Outcomes are shown only after the decision.}
\label{tab:chemlex_program_lineage}
\scriptsize
\setlength{\tabcolsep}{3.7pt}
{\renewcommand{\arraystretch}{1.06}
\begin{tabular}{@{}rrlllllrr@{}}
\toprule
$t$ & $|\mathcal H_t|$ & Suggested$\rightarrow$applied & Program used & Post-reveal program & Selected path & Candidate & $y_t$ & $\bar b_{t+1}$ \\
\midrule
1  & 5  & Regenerate & P1  & P1                    & Reference      & \path|cand_000079| & 11.10 & 91.71 \\
2  & 6  & Revise  & P2  & P1 (restored)         & Reference      & \path|cand_001173| & 46.96 & 91.71 \\
3  & 7  & Revise  & P3a & P2 (restored)         & Reference      & \path|cand_006844| & 0.22  & 91.71 \\
4  & 8  & Revise  & P3b & P3b                   & Alternative C2 & \path|cand_000963| & 93.42 & 93.42 \\
5  & 9  & Retain  & P3b & P3b                   & Reference      & \path|cand_004320| & 1.51  & 93.42 \\
6  & 10 & Revise$\rightarrow$Retain & P3b & P3b            & Alternative C3 & \path|cand_001842| & 94.74 & 94.74 \\
7  & 11 & Retain  & P3b & P3b                   & Reference      & \path|cand_008659| & 0.00  & 94.74 \\
8  & 12 & Revise$\rightarrow$Retain & P1 & P1              & Reference      & \path|cand_010456| & 61.09 & 94.74 \\
9  & 13 & Revise  & P1$^\prime$ & P3b (restored)   & Reference      & \path|cand_008624| & 11.28 & 94.74 \\
10 & 14 & Revise  & P4  & P1$^\prime$ (restored) & Reference    & \path|cand_009455| & 87.56 & 94.74 \\
\bottomrule
\end{tabular}
}
\par\vspace{3pt}
\begin{minipage}{0.97\textwidth}
\scriptsize
P1=\path|public_similarity_reagent_ranker_v1@v001|;
P2=\path|public_similarity_balance_v2@v002|;
P3a=\path|public_similarity_balance_v3@v002|;
P3b=\path|public_similarity_balance_v3@v003|;
P1$^\prime$=\path|public_similarity_reagent_ranker_v2@v002|;
P4=\path|public_component_similarity_balance_v4@v004|.
Version suffixes are artifact versions, not adaptive-round indices.
\end{minipage}
\end{table*}

The following excerpts show how two scoring programs changed after new outcomes were observed. We show changes to scores and reason codes while omitting input parsing, stable sorting, and diagnostic bookkeeping. Each excerpt uses only permitted candidate attributes and revealed outcomes.

\paragraph{\olympus{}: adding local diversity to a rule based on ligand identity and continuous conditions.}
In seed 3, the initial scoring program used the best observed ligand identity together with proximity in the continuous conditions. At round 4, the gate selected M2, the L3 condition with residence time 120.0, temperature 110.0, and catalyst loading 2.515 in Table~\ref{tab:olympus_case_conditions}. Before the outcome was revealed, the trace stored aggregate reference rank 1, support from two distinct rankers, positive gate feature total 0.662, and penalty gate feature total 0.044. After observing its emulated yield, the program-evolution component retained the ligand preference but added penalties for near duplicates and a bounded exploration bonus.

\paragraph{\chemlex{}: balancing positive similarity and avoidance of pairs with low conversion.}
\chemlex{} seed 11 has two selected exploitation alternatives from the same revised scoring program. After low or nonimproving conversions, the program emphasized acid and amine similarity to observations with higher conversion. It also downweighted pairs associated with low conversion while retaining reagent and solvent features. The selected alternatives are C2 and C3 in Table~\ref{tab:chemlex_case_conditions}; Table~\ref{tab:case_audit_records} reports their gate features.

\noindent\begin{minipage}{\dimexpr\linewidth-\parindent\relax}
\noindent\textbf{\olympus{} before revision.}
\begin{lstlisting}[style=appendixcode,language=Python]
score = (0.65 * lig_prior + 10.0 * temp_c +
         8.0 * time_c + 9.0 * load_c +
         lig_bonus + best_onehot_match)
if near_best:
    reason_code = "best_ligand_near_best"
elif lig == best_lig:
    reason_code = "best_ligand_exploration"
else:
    reason_code = "condition_proximity"
\end{lstlisting}
\end{minipage}\par\smallskip

\noindent\begin{minipage}{\dimexpr\linewidth-\parindent\relax}
\noindent\textbf{\olympus{} after the diversity revision.}
\begin{lstlisting}[style=appendixcode,language=Python]
diversity_penalty = (
    5.0 * nearest_same if tested_same_count > 0
    else 1.5 * nearest_any
)
exploration_bonus = (
    2.0 if lig not in seen_ligands else
    1.0 if tested_same_count == 1 else 0.0
)
base = (0.65 * lig_prior + 9.0 * temp_c +
        7.0 * time_c + 8.0 * load_c +
        lig_bonus + best_onehot_match)
score = base + exploration_bonus - diversity_penalty
if near_best and nearest_same < 0.98:
    reason_code = "best_ligand_refined"
elif lig not in seen_ligands:
    reason_code = "new_ligand_exploration"
else:
    reason_code = "condition_proximity_diverse"
\end{lstlisting}
\end{minipage}\par\smallskip

\noindent\begin{minipage}{\dimexpr\linewidth-\parindent\relax}
\noindent\textbf{\chemlex{} before revision.}
\begin{lstlisting}[style=appendixcode,language=Python]
score = (90.0 * learned_similarity +
         28.0 * best_pair_similarity -
         35.0 * neg_penalty +
         reagent_bonus + solvent_bonus +
         8.0 * size_pref + novelty_bonus -
         repeat_penalty)
if pair_key in bad_pair:
    reason_code = "avoids_bad_repeat_pair"
elif learned_similarity > 0.08 or best_pair_similarity > 0.1:
    reason_code = "balanced_positive_similarity"
elif novelty_bonus >= 5.0:
    reason_code = "novel_identity_mix"
else:
    reason_code = "weak_public_signal"
\end{lstlisting}
\end{minipage}\par\smallskip

\noindent\begin{minipage}{\dimexpr\linewidth-\parindent\relax}
\noindent\textbf{\chemlex{} after the balancing revision.}
\begin{lstlisting}[style=appendixcode,language=Python]
score = (100.0 * learned + 30.0 * best_sim -
         25.0 * neg_pen + reagent_bonus +
         solvent_bonus + 6.0 * size_pref +
         novelty - repeat)
if pair in bad_pair:
    reason_code = "avoid_repeat"
elif learned > 0.08 or best_sim > 0.1:
    reason_code = "pattern_match"
elif novelty > 0.0:
    reason_code = "novel"
else:
    reason_code = "baseline"
\end{lstlisting}
\end{minipage}

The \olympus{} revision retains ligand identity and proximity in the continuous conditions while adding diversity and terms that discourage repetition. The \chemlex{} revision increases the weight on positive similarity, reduces the penalties derived from nonpositive observations and repeated low-conversion pairs, and keeps novelty secondary. Literal source names such as \texttt{bad\_pair} and \texttt{avoids\_bad\_repeat\_pair} are preserved from the generated artifact; operationally, they mark previously observed pairs with low conversion, not chemically failed reactions. All excerpts score only permitted candidate attributes and revealed outcomes.

\section{Dataset and Replay Protocol Details}
\label{sec:appendix_protocol}

\paragraph{Datasets.}
\olympusdataset{} contains 5{,}670 candidates. Ligand identity is stored in seven one-hot indicator columns; residence time, temperature, and catalyst loading are additional permitted attributes. The objective is emulated yield (\%). Emulated catalyst turnover number (TON), in moles of product per mole of catalyst, is not supplied to or optimized by any method. \chemlexdataset{} contains 11{,}088 candidates after the prespecified duplicate protocol. Its permitted attributes are the carboxylic acid substrate, amine substrate, reagent mixture, and solvent, and its objective is reported conversion (\%). Each of the five \bhsuite{} tasks contains 792 measured-yield conditions over aryl halide, additive, ligand, and base. \directarylation{} contains 1{,}728 measured-yield conditions over ligand, base, solvent, temperature, and concentration \citep{ahneman2018reaction,shields2021edbo,hickman2023olympusenhanced}.

\begin{center}
\begin{minipage}{\linewidth}
\centering
\captionof{table}{Replay interface used by all methods in the matched suite.}
\label{tab:dataset_replay_interface}
\small
\setlength{\tabcolsep}{2.0pt}
{\renewcommand{\arraystretch}{1.03}
\begin{tabular}{@{}>{\raggedright\arraybackslash}p{0.33\linewidth}>{\raggedright\arraybackslash}p{0.59\linewidth}@{}}
\toprule
Replay item & Fixed value or boundary \\
\midrule
Candidate pools & 5{,}670 for \olympus{}; 11{,}088 for \chemlex{}; 792 for each B--H task; 1{,}728 for \directarylation{}. \\
Permitted attributes & Native reaction variables listed above; every method sees the same task-specific fields and candidate identifiers. \\
Objectives & Emulated yield for \olympus{}, reported conversion for \chemlex{}, and measured yield for B--H A--E and \directarylation{}; unrevealed outcomes remain private to the evaluator. \\
Use of complete benchmark & Complete-pool extrema and ranks are used only after replay for normalized regret, normalized BSF-AUC, Top-1\% Success@15, and normalized plots. \\
Matched protocol & 30 seeds, five initial observations, ten selections, shared candidate pool, evaluator, and deterministic tie-breaking per seed. \\
\bottomrule
\end{tabular}
}
\end{minipage}
\end{center}

\paragraph{Artifact licenses and intended use.}
The benchmark suite supports offline evaluation of optimization methods. The article underlying \olympus{} is distributed under a Creative Commons Attribution 4.0 International license, and the Olympus framework is released for optimization and experiment planning research \citep{sin2025parallel,olympus2021}. The \chemlex{} Zenodo record used for the carboxylic acid and amine coupling replay is distributed under a Creative Commons Attribution Non Commercial 4.0 International license \citep{chemlexzenodo2025,Zhong_2025}. The B--H and \directarylation{} tables are taken from the expanded Olympus collection and attributed to their original HTE studies \citep{hickman2023olympusenhanced,ahneman2018reaction,shields2021edbo}. Our use is limited to noncommercial offline replay and aggregate reporting. We do not redistribute the original archives in the submission or use the benchmarks to recommend or execute live reactions. Outcomes remain unavailable to the controller until a condition is selected.

\paragraph{Objective scales and benchmark maxima.}
The terminal-incumbent and BSF-AUC diagnostics use the objective scales in the released data. \olympus{} emulated yield ranges from 0.0\% to a maximum of 89.2348\%; values near 89 therefore lie near the top of this benchmark rather than a 100\% endpoint. Cleaned \chemlex{} reported conversion ranges from 0.0\% to a maximum of 100.0\%. For every task, normalized regret is the terminal distance from its complete-pool maximum divided by its complete-pool range, and normalized BSF-AUC averages the corresponding range-normalized post-selection incumbents. The component table uses the same task-specific ranges to express differences on a common $[0,100]$ scale. Pool extrema are used only after replay for evaluation. In all percentage-labeled tables, we report 100 times these unit-scale normalized quantities.

\paragraph{\chemlex{} duplicate handling.}
Conditions are grouped by the acid, amine, reagent mixture, and solvent fields. Within each group, reported conversion is averaged after removing groups whose replicate range is at least 50 percentage points. This removes 63 groups containing 126 raw rows and leaves 11{,}088 unique conditions. Candidate identifiers are hashes of the four condition fields, and shuffle seed 0 fixes their replay order.

\paragraph{Matched replay.}
All methods are evaluated with 30 seeds, five initial observations per seed, and 10 sequential selections. For a given seed and task, methods receive the same initial observations, remaining candidate pool, budget, and evaluator. Complete-pool extrema and ranks are used only afterward to compute normalized regret, normalized BSF-AUC, Top-1\% Success@15, and normalized trajectory plots. The Top-1\% set contains $\lceil0.01N\rceil$ candidates---8 for each B--H task and 18 for \directarylation{}---and includes both the five initial observations and ten adaptive selections. Ties are resolved by the recorded \texttt{candidate\_id}.

\section{Prompt Transparency}
\label{sec:appendix_prompts}

This section reproduces the developer messages, instructions, and JSON schemas used by the planner and scoring-program writer. The runtime schema predates the paper terminology, so the prompt artifacts retain their literal names.

\begin{center}
\begin{minipage}{\linewidth}
\centering
\captionof{table}{Mapping from literal prompt terminology to the terminology used in the paper.}
\label{tab:artifact_term_mapping}
\small
\setlength{\tabcolsep}{3.0pt}
{\renewcommand{\arraystretch}{1.05}
\begin{tabular}{@{}>{\ttfamily\raggedright\arraybackslash}p{0.35\linewidth}>{\raggedright\arraybackslash}p{0.57\linewidth}@{}}
\toprule
Literal artifact term & Meaning in this paper \\
\midrule
\path|public|, \path|public_safe| & Information available when a decision is made. \\
\path|skill|, \path|policy| & Scoring program. \\
\path|create_skill| & Regenerate the scoring program. \\
\path|patch_skill| & Revise the scoring program. \\
\path|reuse_active_skill| & Retain the scoring program. \\
\path|rollback_after_reveal| & Post-reveal state update that restores a prior validated scoring program. \\
\path|self_evolving_*|, \path|finite-pool| & Legacy schema identifier; finite candidate set. \\
\path|low-risk|, \path|risk_budget| & Scope of a code edit, unrelated to chemical risk or gate penalties. \\
\path|lower-yield| & Generic lower objective value; the objectives are emulated yield on \olympus{} and conversion on \chemlex{}. \\
\path|minerva_csv| & Literal dataset identifier for \olympusdataset{}. \\
\path|chemlex_csv| & Literal dataset identifier for \chemlexdataset{}. \\
\path|macro_frontier_scout| & Exploration refinement. \\
\path|residual_scout| & Exploitation refinement. \\
\bottomrule
\end{tabular}
}
\end{minipage}
\end{center}

Runtime prompts contain revealed observations, summaries of permitted candidate attributes, recorded controller state, memory, and earlier gate reports. They exclude unrevealed outcomes, ranks over the complete benchmark, and evaluator-only state. Revision requests additionally supply the current scoring-program source through the literal field \texttt{active\_skill\_source}.

\begin{center}
\begin{minipage}{\linewidth}
\centering
\captionof{table}{Prompt interface for the two language model components. Runtime values are decision-time summaries or scoring-program source strings.}
\label{tab:prompt_interface}
\small
\setlength{\tabcolsep}{5.0pt}
{\renewcommand{\arraystretch}{1.12}
\begin{tabular}{@{}>{\raggedright\arraybackslash}p{0.16\linewidth}>{\raggedright\arraybackslash}p{0.35\linewidth}>{\raggedright\arraybackslash}p{0.39\linewidth}@{}}
\toprule
Prompt & Inputs available at decision time & Required output \\
\midrule
Planner & Observed summary, candidate summary, memory summary, controller state, objective-value trend, last gate report, parser error if any. & One structured JSON task plan choosing \path|create_skill|, \path|patch_skill|, or \path|reuse_active_skill|. \\
Scoring program synthesis & Task plan, recent revealed examples, candidate schema and summary, memory summary, controller state, active program source if revising, and any parser error. & One structured JSON artifact defining \texttt{rank\_candidates} over revealed observations and permitted candidate attributes. \\
\bottomrule
\end{tabular}
}
\end{minipage}
\end{center}

\subsection{Planner Prompt Card}

\begin{center}
\begin{minipage}{\linewidth}
\centering
\captionof{table}{Complete input fields for the planner prompt. Literal field names are unchanged.}
\label{tab:planner_prompt_inputs}
\small
\setlength{\tabcolsep}{3.0pt}
{\renewcommand{\arraystretch}{1.02}
\begin{tabular}{@{}>{\ttfamily\raggedright\arraybackslash}p{0.31\linewidth}>{\raggedright\arraybackslash}p{0.61\linewidth}@{}}
\toprule
Field & Contents \\
\midrule
task & \path|plan_next_public_safe_policy_edit|. \\
\path|schema_version|, \path|run_id|, \path|round_index| & Replay identifiers and planner schema version. \\
\path|observed_summary| & Summary of revealed observations only. \\
\path|candidate_summary| & Summary of the remaining candidate pool and permitted attributes. \\
\path|memory_summary| & Memory text truncated to 1200 characters. \\
\path|policy_state| & Compact controller state, including active program identifiers and revealed histories. \\
\path|reward_trend| & Trend computed from revealed objective values. \\
\path|last_gate_report| & Compact report from the previous deployment/intervention gate. \\
\path|parser_error| & Parse or validation error from a previous retry, truncated to 500 characters. \\
\path|allowed_actions| & \texttt{create\_skill}, \texttt{patch\_skill}, \texttt{reuse\_active\_skill}. \\
\path|allowed_skill_families| & \texttt{ranker}, \texttt{constraint}, \texttt{exploration}, \texttt{calibrator}, \texttt{fallback}. \\
\bottomrule
\end{tabular}
}
\end{minipage}
\end{center}

\noindent\textbf{Developer message.}\par
\begin{lstlisting}[style=appendixprompt,basicstyle=\scriptsize\ttfamily]
You are the public-safe planner for an offline finite-pool optimization harness. Return only the JSON object required by the schema. Use only public observed rows, public candidate features, memory, and gate/reward summaries supplied by the user.
\end{lstlisting}

\noindent\textbf{Planner instructions.}\par
{\small
\begin{enumerate}
\item Plan only one next edit/action for a language-generated optimizer skill.
\item Use only public views and revealed objective values; do not seek evaluator-private labels, answer keys, baseline artifacts, files, network, or credentials.
\item Prefer low-risk deterministic code edits that compile to \texttt{rank\_candidates}.
\item The final deployed tool must score the full remaining \texttt{candidate\_df}, not a menu.
\item Reuse is allowed only when the active skill is both deployable and still improving or diversifying selected objective values.
\item Do not patch solely because one round failed after a recent best-observed improvement; reuse once unless the last selected objective value was clearly poor.
\item Plan \texttt{patch\_skill} mainly after two or more consecutive no-improvement rounds, or after a revealed selection far below the current best.
\item If patching after stagnation, blend the active scoring rule with bounded diversity/novelty; do not replace a previously improving policy with an unrelated scorer.
\item If the active skill repeatedly selects lower-yield near-duplicates after a first improvement, patch it to avoid over-exploiting that local region.
\item Return JSON with \texttt{task\_plan} and \texttt{self\_reported\_forbidden\_info\_used=false}.
\end{enumerate}
}

\noindent\textbf{Planner output schema.}\par
\begin{lstlisting}[style=appendixprompt]
{"schema_version":"self_evolving_task_plan_v1","run_id":"[RUN_ID]","round_index":"[ROUND_INDEX]",
 "task_plan":{"action":"create_skill|patch_skill|reuse_active_skill","skill_family":"ranker|constraint|exploration|calibrator|fallback",
 "objective":"short public-safe objective","target_skill_id":"active skill id or null","risk_budget":"low|medium|high",
 "required_checks":"[DEFAULT_REQUIRED_CHECKS]","rationale":"short public-safe rationale"},
 "self_reported_forbidden_info_used":false}
\end{lstlisting}

\subsection{Scoring Program Synthesis Prompt Card}

\noindent\textbf{Developer message.}\par
\begin{lstlisting}[style=appendixprompt]
You are the public-safe skill author for an offline finite-pool optimization harness. Return only the JSON object required by the schema. The source must define exactly rank_candidates(observed_df, candidate_df, memory=None, tool_state=None). Do not read files, call networks, inspect DataFrame attrs, or use evaluator-private outcomes.
\end{lstlisting}

\begin{center}
\begin{minipage}{\linewidth}
\centering
\captionof{table}{Complete input fields for the scoring-program synthesis prompt. Literal field names are unchanged.}
\label{tab:scoring_program_prompt_inputs}
\small
\setlength{\tabcolsep}{3.0pt}
{\renewcommand{\arraystretch}{1.02}
\begin{tabular}{@{}>{\ttfamily\raggedright\arraybackslash}p{0.31\linewidth}>{\raggedright\arraybackslash}p{0.61\linewidth}@{}}
\toprule
Field & Contents \\
\midrule
task & \path|generate_or_patch_public_safe_rank_candidates_skill|. \\
\path|schema_version|, \path|run_id|, \path|round_index| & Replay identifiers and program synthesis schema version. \\
\path|task_plan| & Planner output for the current round. \\
\path|observed_examples| & Up to the last 12 revealed rows, with cells compacted for prompt length. \\
\path|candidate_schema| & Permitted candidate columns and data types. \\
\path|candidate_summary| & Remaining candidate count, permitted columns, and categorical cardinalities. \\
\path|memory_summary| & Memory text truncated to 1200 characters. \\
\path|policy_state_summary| & Active program identifiers plus recent outcome and gate histories. \\
\path|active_skill_source| & Current scoring program source, truncated to 8000 characters when revising; empty otherwise. \\
\path|parser_error| & Parse or validation error from a previous retry, truncated to 500 characters. \\
\bottomrule
\end{tabular}
}
\end{minipage}
\end{center}

\noindent\textbf{Scoring program synthesis instructions.}\par
{\small
\begin{enumerate}
\item Write normal Python source defining exactly \texttt{rank\_candidates(observed\_df, candidate\_df, memory=None, tool\_state=None)}.
\item Return a dictionary, never a DataFrame: \texttt{\{'ranked\_candidates': list, 'tool\_state': dict, 'tool\_diagnostics': dict\}}.
\item Each ranked row must include \texttt{candidate\_id}, \texttt{rank}, \texttt{score}, \texttt{reason\_code}, and \texttt{evidence\_refs}. It must cover every \texttt{candidate\_df} row with unique positive ranks and finite scores.
\item \texttt{evidence\_refs} is empty unless exact \texttt{observation\_id} values are copied from \texttt{observed\_df}; never invent evidence strings.
\item Use only public \texttt{observed\_y} for observed rows and public candidate features.
\item Do not read files, call networks, import disallowed modules, inspect DataFrame attrs, or reference evaluator-private labels, answer keys, baseline artifacts, private provenance fields, or credentials.
\item Do not use double-underscore names or strings in temporary columns, helper variables, imports, or escape hatches.
\item Do not call \texttt{getattr}, \texttt{setattr}, \texttt{hasattr}, \texttt{eval}, \texttt{exec}, \texttt{compile}, \texttt{globals}, \texttt{locals}, \texttt{vars}, \texttt{open}, or \texttt{\_\_import\_\_}.
\item Make ranking row-order invariant: do not use enumerate index, original row order, DataFrame index, or insertion order in scores or tie-breaks.
\item Sort with public score first and \texttt{candidate\_id} as the only deterministic tie-breaker; never tie-break by row position.
\item Avoid creating helper columns such as \texttt{\_row\_order}, \texttt{\_index}, \texttt{\_\_lig}, or \texttt{\_position} for ordering.
\item Return JSON with \texttt{skill\_artifact.source} and \texttt{self\_reported\_forbidden\_info\_used=false}.
\end{enumerate}
}

\noindent\textbf{Scoring program synthesis output schema.}\par
\begin{lstlisting}[style=appendixprompt]
{"schema_version":"self_evolving_skill_artifact_v1","run_id":"[RUN_ID]","round_index":"[ROUND_INDEX]",
 "skill_artifact":{"skill_id":"short id","family":"[TASK_PLAN_SKILL_FAMILY]","source":"Python source string defining rank_candidates","rationale":"short public-safe rationale"},
 "self_reported_forbidden_info_used":false}
\end{lstlisting}